\RequirePackage{fix-cm}

\documentclass[twocolumn]{svjour3}

\usepackage{cite}
\usepackage{amsmath,amssymb,amsfonts}
\usepackage{algorithmic}
\usepackage{graphicx}
\usepackage{subfig}
\usepackage{textcomp}
\usepackage{xcolor}
\def\BibTeX{{\rm B\kern-.05em{\sc i\kern-.025em b}\kern-.08em
    T\kern-.1667em\lower.7ex\hbox{E}\kern-.125emX}}
\usepackage{enumitem}
\usepackage{soul}
\usepackage{hyperref}

\begin{document}\sloppy

\title{\LARGE \bf Deployment and Evaluation of a Flexible Human-Robot Collaboration Model Based on AND/OR Graphs in a Manufacturing Environment
}

\author{
Prajval Kumar Murali, Kourosh Darvish$^{1}$, Fulvio Mastrogiovanni
\thanks{
$^{1}$Corresponding author's email: kourosh.darvish@gmail.com
}}

\institute{
All the authors are with the Department of Informatics, Bioengineering, Robotics, and Systems Engineering, University of Genoa, Via Opera Pia 13, 16145, Genoa, Italy.
}

\maketitle

\thispagestyle{plain}
\pagestyle{plain}

\begin{abstract}
The Industry 4.0 paradigm promises shorter development times, increased ergonomy, higher flexibility, and resource efficiency in manufacturing environments.
Collaborative robots are an important tangible technology for implementing such a paradigm.
A major bottleneck to effectively deploy collaborative robots to manufacturing industries is developing \textit{task planning} algorithms that enable them to recognize and naturally adapt to varying and even unpredictable human actions while simultaneously ensuring an overall efficiency in terms of production cycle time.
In this context, an architecture encompassing task representation, task planning, sensing, and robot control has been designed, developed and evaluated in a real industrial environment.
A pick-and-place palletization task, which requires the collaboration between humans and robots, is investigated.
The architecture uses AND/OR graphs for representing and reasoning upon human-robot collaboration models online.
Furthermore, objective measures of the overall computational performance and subjective measures of naturalness in human-robot collaboration have been evaluated by performing experiments with production-line operators.
The results of this user study demonstrate how human-robot collaboration models like the one we propose can leverage the flexibility and the comfort of operators in the workplace. In this regard, an extensive comparison study among recent models has been carried out.
\end{abstract}


\section{Introduction}
\label{Sec:introduction}
Robots have been used in manufacturing industries from the late 1960s.
Industrial robots are very efficient when performing repetitive tasks with high accuracy, and are ideally suited for high volume and low mix production.
Between 2018 and 2021, it is estimated that almost 2.1 million new industrial robots are being installed in factories around the world \cite{ifr:1}. 
However, most of the industrial robots are physically separated from human workers by fences and tasks are organized such that to have separate workspaces for humans and robots. 
These limits pave the way for the introduction of collaborative robots.

\begin{figure}[!t]
    \centering
    \includegraphics[width=1.0\columnwidth]{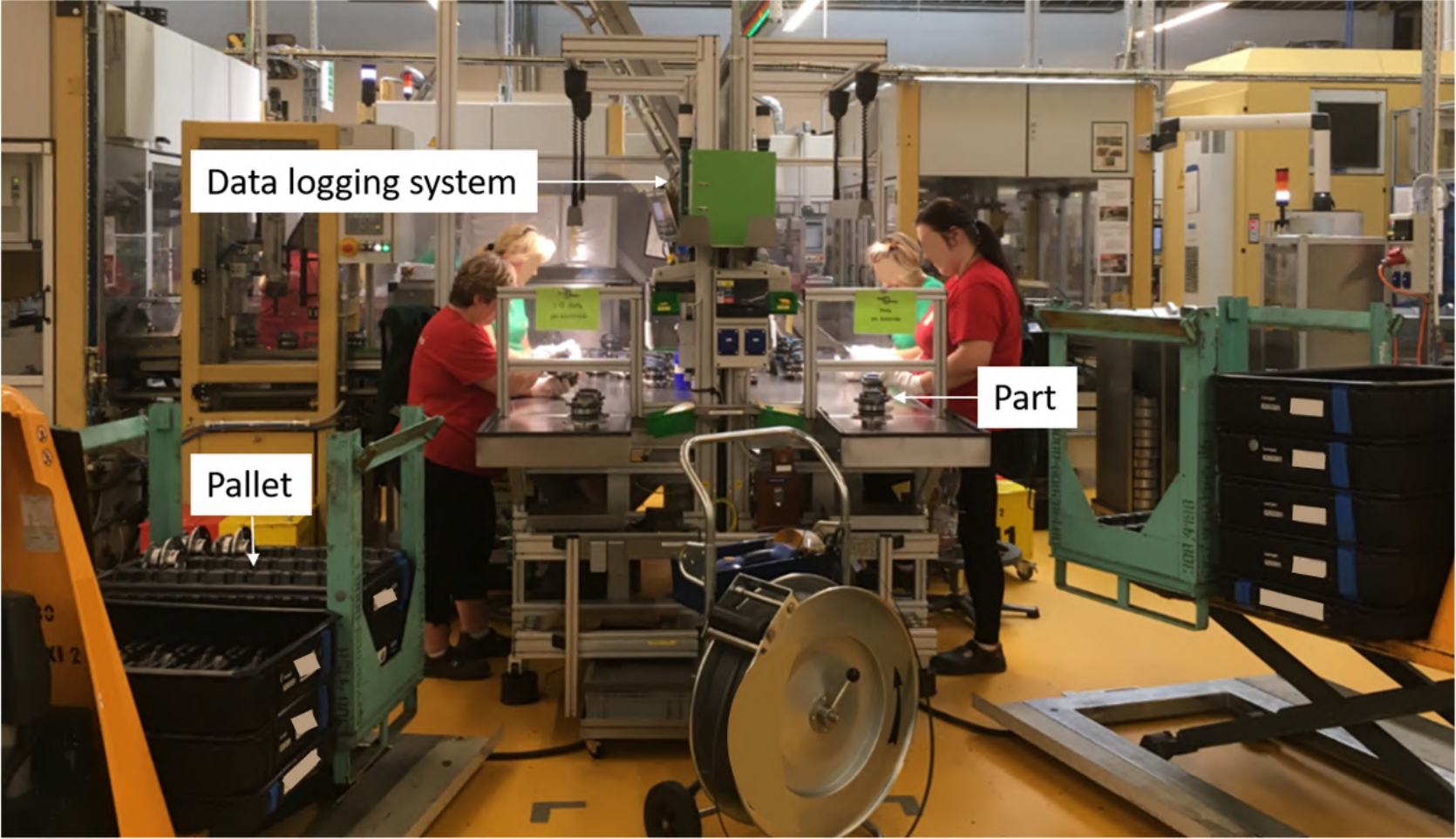}
    \caption{The target application at a Schaeffler plant with production-line operators performing part inspection and palletization (image courtesy of Schaeffler Group).}
    \label{fig:application_process}
\end{figure}

A \textit{collaborative robot} or \textit{cobot} is a robot intended to interact with humans in a shared workspace \cite{colgate1996cobots}. 
Cobots can act as co-workers alongside humans in many applications such as manufacturing assembly lines, or logistics.
Cobots in the literature date back to pioneering work described in~\cite{akella1999cobots} as a result of a General Motors backed initiative to find a way to make robots or robot-like equipment safe enough to team with people.
In manufacturing industries, cobots can reduce the physical and cognitive stress of human operators in the assembly line, and simultaneously improve quality, productivity and safety~\cite{bicchi2008safety}. 
This is a key issue, since according to statistics from the Occupational Safety and Health Department of the US Department of Labour, more than 30\% of European manufacturing operators are affected by lower back pain, leading to enormous social and economic costs~\cite{cherubini2016collaborative}.
Hence, ergonomy is a major motivating factor for studying human-robot collaboration (HRC) from an industrial point-of-view.
Furthermore, with collaborative robots we combine the best of both worlds: precision, speed and efficiency of robots with human-like cognitive capabilities and dexterity in dynamic environments.

Deploying human-robot collaboration in industrial assembly tasks is a challenging problem as human operators can introduce potential non-determinism by their actions as the task progresses. 
Robots need to be able to plan complex sequences of actions, which involve collaboration and common goals with human operators or other robots~\cite{johannsmeier2017hierarchical}.
Furthermore, an optimal task sharing between the human and the robot counterparts is essential for collaboration.
Ideally, when a certain freedom in task allocation is possible, it has been argued that human satisfaction levels are higher when they have the freedom to choose the tasks to perform~\cite{munzer2017impact}.
Consequently, an optimal and efficient task representation, task allocation and task planning framework is of the utmost importance.
Furthermore, in order to enforce an effective collaboration, human operators need to \textit{trust} their robot partners to operate in a safe and efficient manner.
There is apprehension among workers in the industry to allow for workspace and task sharing as swift and unpredictable robot motions in close proximity generate fear in human operators~\cite{arai2010assessment}.
Hence, the need arises to perform experimental studies whereby factory workers operate in close proximity to robots, which may also involve physical interaction \cite{Capitanellietal2018, Cannataetal2010}.
These aspects represent the main motivations for this paper.

The application targeted here is the collaborative human-robot palletization of automotive wheel-bearing parts, as shown in Figure \ref{fig:application_process}.
Pallets are stacked one over another into a packaging box.
Human operators receive the part on a conveyor from the previous stage of the production line.
Then, they perform tactile inspection on both sides of the part to detect sharp edges or burrs, as well as visual inspection for scratches and defects on the polished surface.
In case of presence of oil or dirt, they clean the part before placing it inside the pallet.
Once the pallet is complete, information regarding part number, date, and time are logged and placed in the pallet, another pallet is placed above the previous one, and the process is repeated.
Human errors, such as missing defects during visual or tactile inspection or forgetting to correctly log details can happen due to enduring physical and mental stress.
More importantly, the repetitive bend-pick-place sequence can cause fatigue conditions \cite{dolan1998repetitive}.
This situation may also result in gender bias, as it might be prohibitive to employ women for such roles.
For these reasons, a collaborative robot is proposed to perform the palletization task after human operators inspect the part, i.e., they conduct part inspection and place it to a designated pickup location.
The collaborative robot, equipped with a gripper and a vision system, recognizes the part and performs the palletization.
For various reasons, during robot motion, human operators must be able to stop the robot by force-contact and to perform the palleting task themselves.

A flexible task allocation and planning framework to properly allocate tasks between humans and robots, as well as to sequence them, is necessary.
This leads to a more precise definition of the functional objectives of this paper:
\begin{enumerate}[label=$\textit{F}_{\arabic*}$]
    \item Design a modular and adaptive software architecture for collaborative robots, which attempts to optimize the following performance metrics\label{obj1}: 
    \begin{enumerate}
        \item \textit{Process-centered}: The architecture must attempt at minimizing the overall cycle time, i.e., the total time from the beginning to the end of the process.
        \item \textit{Human-centered}: The architecture must be flexible, i.e., human operators must not be forced to carry out a strictly predefined sequence of actions, but should be allowed to choose an action to perform \textit{online}, whereas the collaborative robot should be able to adapt \textit{reactively}. It should also ensure ergonomy so that human operators avoid the picking and placing of heavy parts.
    \end{enumerate}
    
    \item \label{obj2} Incorporate a pseudo-linguistic or symbolic communication level between human operators and robots, such that it enables a bi-directional and intuitive interaction.

    \item \label{obj3} Study the trade-off between the expressivity of the employed formalisms and the associated computational performance, as the system is expected to support human-robot collaboration in a real production line.
\end{enumerate}

This paper is organized as follows.
Section \ref{sec:background} summarizes the state-of-art in task planning for human-robot collaborative manufacturing and highlights our contributions in the context of current and related research.
The proposed architecture and the theoretical formalism regarding task representation and task planning are presented in Section \ref{sec:architecure}.
Experimental results are reported in Section \ref{sec:experiment}, and finally summarized in the Conclusions (Section \ref{sec:conclusion}). 
\section{Background}
\label{sec:background}
The work presented in this paper touches upon different aspects of HRC architectures, namely task representation, task planning and task allocation integrated with motion planning and natural interfaces for human-robot collaboration.
In this Section, we discuss relevant approaches discussed in the literature.

In the paragraphs that follow and throughout the paper, we refer to human operators or robots indifferently as \textit{agents}.
In human-robot collaboration scenarios, all the involved agents can have a shared plan to achieve a common goal.
Hence, it is necessary for the agents to recognize the intentions of the other agents involved in the cooperation.
The objective of task allocation in HRC is to suitably distribute the tasks to be carried out among the robot and the human operator while considering their capabilities in order to improve work quality~\cite{ranz2017capability}.
Such a problem has been investigated in various scenarios such as manufacturing, healthcare, space missions and in the military sector \cite{pintoreview}.
The impact of task allocation on the performance of human operators has been studied by Gombolay \textit{et al.} \cite{gombolay2015coordination}.
They argue that the highest human satisfaction levels are reached when human operators have the freedom to choose the tasks, and the overall performance of the team is high.
Results show that human operators prefer autonomous task allocation \textit{as provided by the robot} while at the same time retaining some level of partial authority over the collaboration process.
Hence, in a HRC scenario, it is important to understand that the robot must ensure an efficient task allocation, and simultaneously allowing human operators to exert a freedom of choice.
In the literature, the task allocation problem has been addressed as a multi-objective optimization problem with constrained resources \cite{chen2011assembly}, \cite{tsarouchi2017human}, \cite{takata2011human}.
These stochastic approaches optimize certain metrics such as the cycle time, human- or robot-related costs, idle time as well as the human cognitive load \cite{johannsmeier2017hierarchical}, \cite{darvish2018interleaved}, \cite{gerkey2004formal}.
For instance, in \cite{takata2011human} a multi-objective optimization formulation based on production volumes is used, and the allocation is based on the minimum expected total production cost calculated through demand forecasting techniques.
Similarly, in \cite{gombolay2016apprenticeship} the problem is modelled as a mixed integer linear programming (MILP) formulation to produce feasible schedules satisfying temporal and spatial constraints.
The approaches in \cite{johannsmeier2017hierarchical}, \cite{tsarouchi2017human}, \cite{darwish2017flexible} are performed offline, whereas those in \cite{chen2014optimal}, \cite{darvish2018interleaved} are executed online.
As observed in \cite{pintoreview}, a robust and dynamic task allocation, claimed to be scalable to large assembly processes, is limited to research and still under-development.
In this paper, to optimize for run-time performance, we have used a fixed task allocation solution based on the capabilities of each agent.    

In a goal-oriented cooperation task, it has been argued that there is greater acceptance and better task performance among human operators when they can understand, explain, and predict robot behaviors \cite{bortot2013directly}.
Therefore, it is necessary to model the peculiar sequence of actions performed by human operators online, and provide robots with the capability to reactively adapt to operator actions \cite{levine2014concurrent}.
These peculiar sequences of tasks or actions can be represented as graph- or tree-like structures.
The recognition of operator actions can be performed upon completion, which incurs delays in the process or can be predicted online for the sake of efficiency.
However, the latter approach may cause disruption in the process in case of mispredictions.
Markov Decision Processes (MDPs) have been used in \cite{crandall2018cooperating}, \cite{toussaint2016relational} to enforce adaptation to human workers online.
However, among different task representation approaches, those based on AND/OR graphs and Task Networks (TNs) explicitly consider efficiency and predictability at the representation level \cite{darvish2018interleaved}, ~\cite{lamon2019capability}.
Behavior Trees (BTs) are another popular approach for task representation.
BTs are a very efficient way to develop software architectures for robots, which are both modular and reactive \cite{colledanchise2018behavior}, \cite{Coronadoetal2018a}, \cite{Coronadoetal2018b}, \cite{MastrogiovanniSgorbissa2013}. 
For instance, CoSTAR \cite{paxton2017costar} is a research project exploring BTs with collaborative robots for HRC.
The use of BTs is intended towards the execution of general-purpose actions that may be learned or planned.
However, BT engines are complex to implement and, in general, they do not offer much advantages over other control architectures when the robot is operating in a structured environment~\cite{colledanchise2017behavior}. 
In this paper, we develop on previous work described in \cite{darwish2017flexible} and \cite{darvish2018interleaved} to represent the cooperation task as an AND/OR graph.

For seamless integration of human operator and robot actions to be performed in a well-defined temporal sequence, a reliable and efficient task planner is crucial.
A vast number of studies have been carried out to investigate the role of task planning in HRC \cite{darwish2017flexible}, \cite{johannsmeier2017hierarchical}, \cite{korsah2013comprehensive}, \cite{hawkins2014anticipating}, \cite{rozo2016exploiting}, \cite{koppula2016anticipating}, \cite{caccavale2017flexible}, \cite{haigh1998planning}\cite{Capitanellietal2018}.
As noted by a recent review on task planning in HRC by Pinto \textit{et al.} \cite{pintoreview}, we can distinguish between three major abstraction levels: system level, team level and agent level.
At the system level, teams of human operators and robots are formed, and the overall sequence of tasks and resources are distributed among them.
At the team level, the operator and the robot must collaborate with each other to complete a given task.
Finally, the agent level planner maps the team level planning to a level of abstraction above the hardware (and therefore directly addressing robot sensors and actuators).
The planning strategies can either take into account certain preferences from human operators in order to provide a suitable sequence of optimal actions \textit{online} leading to the cooperation goal \cite{wilcox2013optimization}, \cite{agostini2011integrating}, \cite{koppula2016anticipating}, \cite{caccavale2017flexible}, \cite{darvish2018interleaved} or \textit{offline} \cite{johannsmeier2017hierarchical}, \cite{hawkins2014anticipating}, \cite{sanderson1988task}.
For the latter case, Johannsmeier and Haddadin \cite{johannsmeier2017hierarchical} propose a multi-agent hierarchical human-robot team approach.
The tasks are modelled using an AND/OR graph.
The team level planner produces task sequences for every agent via an $A^*$ graph search induced by the AND/OR graph.
Each agent implements its modular skills via hierarchical and concurrent state machines in order to map abstract task descriptions to the subsequent real-time level.
However, a noticeable limitation of that approach is that only the agent level (responsible for real time operations) is online, whereas the rest of the process is offline, i.e., it is a \textit{static} architecture based on AND/OR graphs.
Similarly, Hawkins \textit{et al.} \cite{hawkins2014anticipating} propose a fixed-task duration AND/OR graph with the goal of anticipating collaborative actions in presence of uncertain sensing or task ambiguity.
The interaction can be improved if human preferences are known or can be inferred, e.g., through direct communication or learned through experience.
When preferences are known \textit{a priori}, plans can be formulated as an optimization problem.
Wilcox \textit{et al.} \cite{wilcox2013optimization} devise an algorithm that takes as input a Simple Temporal Problem with Preferences (STPP), which encloses the variables, constraints, preferences and the optimization function.
The outcome is a dispatchably optimal form of the STPP, where each executable event is assigned with a time within specified time bounds and the preference function is maximized.
As anticipated above, instead of coding human preferences beforehand, they can also be learnt from experience.
Agostini \textit{et al.} \cite{agostini2011integrating} approach the problem by using a STRIPS-like planner \cite{fikes1971strips} that is constantly refined by a decision-making algorithm.
It is up to the human to play the role of supervisor, in the case of failure in the planning decisions. 
Similarly, Markov models are used in \cite{rozo2016exploiting}, \cite{koppula2016anticipating} to model human and robot behaviors with the goal of anticipating human actions to complete the task.
Finally, task allocation and planning can be modified online by acquiring human preferences via speech, gestures or other user interfaces.
In \cite{caccavale2017flexible}, the authors propose a hierarchical task planner that can be adapted online through speech and gestures.
These signals are fused to recognize human intentions on the fly, and therefore they are suitable for action replanning.
Furthermore, in \cite{darwish2017flexible} and \cite{darvish2018interleaved} a framework representing the tasks as AND/OR graphs allowing for \textit{online} adaptation is presented.
This enables a human operator to choose freely among a number of alternatives to perform a task.
Human preferences are acquired through gesture recognition via wearable devices \cite{Brunoetal2014, Carfietal2018}.
This framework was validated using a human operator and a Baxter-dual arm manipulator performing turn taking actions in any allowed sequence.

\section{System's Architecture}
\label{sec:architecure}

\subsection{Rationale}

\begin{figure}[!t]
    \centering
    \includegraphics[width=1.0\columnwidth]{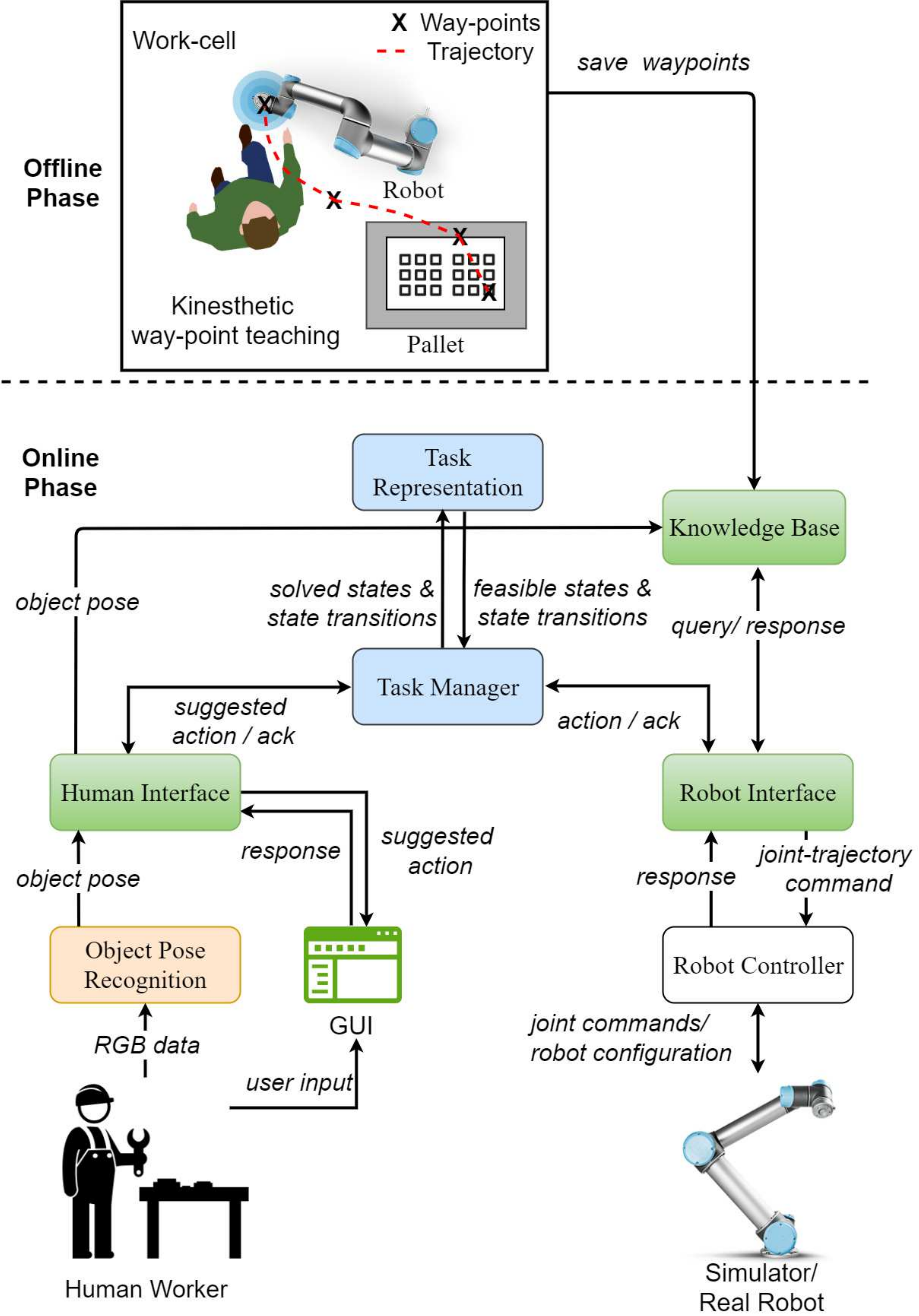}
    \caption{A graphical description of the overall architecture showing the various modules and data flow.}
    \label{fig:arch}
\end{figure}

The overall structure of the system we propose is based on a modular, hybrid reactive-deliberative architecture, which is conceptually sketched in Figure \ref{fig:arch}.
The workflow is organised into two phases, namely the \textit{offline} phase and the \textit{online} phase.
The \textit{offline} phase is related to the teaching of safe way points for the robot in the manufacturing work-cell.
This process is application-dependent and may be done by a robot programmer using the robot teach-pendant \cite{krause2003robotic}, \textit{Programming by Demonstration} (PbD) \cite{billard2004robot} or \textit{Offline Programming} (through CAD-based simulations) \cite{hoener1985offline}, or even using \textit{End User Development} (EUD) techniques \cite{Coronadoetal2018a}.
In this paper, we use PbD techniques, in particular \textit{kinesthetic teaching} wherein the robot programmer physically guides the robot in performing the skill \cite{akgun2012trajectories}, \cite{Carfietal2018}.
However, other approaches would be equally legitimate.

The \textit{online} phase has three layers, namely the \textit{representation layer} shown in blue in Figure~\ref{fig:arch}, the \textit{perception layer} in orange and the \textit{action layer} in green.
The representation layer maintains all the necessary information regarding the overall cooperation process in the \textit{Task Representation} module.
It is also responsible for task allocation, planning and execution via the \textit{Task Manager} module.
The perception layer, comprising of the \textit{Object Pose Recognition} module, recognizes objects to grasp in the workspace and provides the object pose to the interface layer. 
The action layer consists of the \textit{Human Interface}, the \textit{Robot Interface} and the \textit{Knowledge Base} modules.
The \textit{Human Interface} is responsible for suggesting actions to the human operator via a \textit{Graphical User Interface} (GUI) and waiting for the response from either the perception layer or the GUI. 
Similarly, the \textit{Robot Interface} generates joint trajectories and serializes the execution of the commanded robot actions to the \textit{Robot Controller}.
The latter sends low-level joint trajectory commands to the robot and continuously monitors joint states.
The architecture reliably handles messages between modules and the interaction between them is necessary for the overall operation.

The layered framework serves as a way of abstracting the complexity at the lowest level of robot action execution from human operator actions via the representation layer.
The \textit{Task Representation} and \textit{Task Manager} modules are extended from the previous work on the FlexHRC framework by Darvish \textit{et al.}~\cite{darwish2017flexible}.
The \textit{Task Representation} module maintains knowledge about all possible states and the transitions between states modelling cooperative tasks.
The module also defines how an HRC process can progress by providing suggestions to the human operator or the robot about the next action to carry out. 
Cooperation models are represented using AND/OR graphs \cite{de1990and}, \cite{darwish2017flexible}, as described in Section~\ref{TRSection}.
The \textit{Task Manager} operates on the graph via an \textit{ad hoc} online traversal procedure to determine the most appropriate sequence of actions to ground the cooperation on, based on the graph structure.
In order to generate the next optimal action, the \textit{Task Manager} must be informed when the previous action is completed successfully or unsuccessfully by the action layer.
However, there is an important difference between the two loops involving \textit{Task Manager} with \textit{Human Interface} and \textit{Robot Interface}, respectively.
In the first case, the \textit{Task Manager} merely suggests the next action to the human operator, leaving them free to execute it or not, therefore taking into account objective \ref{obj1}, whereas in the second case it imposes the next action for the robot to perform.
Furthermore, the \textit{Task Manager} does not have access to low level robot trajectory planning and internal parameters in the \textit{Robot Interface}, as we employ a modular component-based software architecture using the \textit{Robot Operating System} (ROS) as our middleware (objective \ref{obj1}) \cite{quigley2009ros}.
As a consequence, the representation layer is \textit{robot independent}, i.e., it can be used along with any robot \textit{capable} of performing the desired actions.
It is noteworthy that this structure may lead to a number of issues as highlighted in \cite{Capitanellietal2018}. 

The \textit{Robot Interface} maps semantic action commands issued by the \textit{Task Manager} to joint action commands for the \textit{Robot Controller} module.
The main functions of the \textit{Robot Interface} are robot motion planning and robot manipulation.
Motion planning is the method of breaking down a desired motion task into discrete movements, which satisfy movement constraints and possibly optimize some aspects of the movement \cite{latombe2012robot}.
In order to move the robot arm to a given position, the individual joints or the position of the end-effector must be controlled.
In our application, the robot has a proprietary internal controller, which accepts either joint trajectory set-points $q$~\cite{andersen2015optimizing} or
joint velocity set-points $\Dot{q}$.
As anticipated above, safe waypoints between the grasp and the goal locations are taught to the robot by an expert through kinesthetic teaching \cite{akgun2012trajectories} in the offline phase and stored as joint positions in the \textit{Knowledge Base}.
The \textit{Robot Interface} chooses the joint positions corresponding to the semantic action commands from the \textit{Knowledge Base} and sends these as target positions to the \textit{Robot Controller}.    
The \textit{Robot Controller} reads the current robot pose and interpolates poses between the current pose and the target pose using a cubic trajectory generator and streams these poses to the robot~\cite{andersen2015optimizing}.
It is important to note that the \textit{Robot Controller} shown in Figure \ref{fig:arch} is in fact a ROS-based driver interacting with the internal robot controller provided by Universal Robots~\cite{URmanual}.
Safe robot trajectories are crucial in an industrial environment and simulation provides robot programmers with fast feedback on the validity of generated robot motions. 
Hence, robot programmers can teach way-points to the robot and test the overall architecture in simulation before deploying to the real robot. 
This is explained in Section \ref{sec:experiment}.

The \textit{Human Interface} handles the information exchange between human operators and robots in a \textit{pseudo multi-modal} approach. 
The \textit{Task Manager} provides the suggested action for the human operator to the \textit{Human Interface}, and the latter provides information regarding the actions through simple messages displayed on a GUI.
It has been shown that GUIs provide the most effective, albeit less natural, method of information exchange between agents in factory settings (objective \ref{obj2}) \cite{goodrich2008human}.
The \textit{Human Interface} module also interacts with the perception layer to detect the conclusion of an operator action.

The perception layer consists of the \textit{Object Pose Recognition} module.
When a human operator delivers the part to the robot after inspection, the part pose (i.e., position and orientation) in world/robot frame needs to be recognized.
The task of the \textit{Object Pose Recognition} module is carried out through feature extraction and 2D pose estimation.
We have a closed world assumption, i.e., when the object pose estimation is completed and the robot has not done any action, it implies that human operators have performed the action themselves.
This information is passed to the \textit{Task Manager} through the \textit{Human Interface} module.
Using well-known hand-eye calibration techniques with fiducial calibration grid~\cite{tsai1989new}, we transform the part pose in camera frame to the robot frame as 
\begin{equation}
    {}^{robot} H_{part} = {}^{robot} H_{tool} . {}^{tool} H_{camera} . {}^{camera} H_{part},
\end{equation}
where ${}^{A}H_{B}$ refers to the $4\times4$ homogeneous transform that multiplied for a point expressed in the frame $B$ transforms it in a point expressed in frame $A$~\cite{traversaroMult}. 
The ${}^{robot} H_{part}$ is stored in the \textit{Knowledge Base}.

Additionally, the architecture also allows human operators to physically collaborate and stop the robot while it is performing the palletization of the part.
Once the robot is stopped, operators can interact with the \textit{Human Interface} module through the GUI to retrieve the part from the robot and perform the palletization themselves, henceforth termed as the \textit{intervention} procedure.  
A change in the operator action is recognised by the robot and the \textit{Task Manager} reactively adapts by commanding the robot to perform a new action.
Hence, human operators are not constrained to perform a strictly pre-defined sequence of actions, therefore addressing objective~\ref{obj1}(b).
The \textit{intervention} process is further elaborated in Section \ref{sec:experiment}.

Given the flexibility and reactive nature of our framework, it must be re-iterated that the architecture assumes that human operators are not trying to \textit{cheat} the system.
Although they are given the freedom to perform an action or not, the system cannot determine whether human operators perform an action completely unrelated to the overall cooperation goal, since we are not actively monitoring them.
Admittedly, the system is not completely fool-proof, principally due to the avoidance of using human activity recognition algorithms \cite{Kareemetal2018}.
Another (practical) assumption worth mentioning is that the pallet pose is fixed and known \textit{a priori}, thus allowing to program robot goal poses \textit{offline}.


\subsection{Task Representation and Task Manager}
\label{TRSection}
As anticipated in the previous Section, the \textit{Task Representation} module encodes the cooperation task as an AND/OR graph.
In general, an AND/OR graph is a logical representation of the reduction of problems (or goals) to conjunctions and disjunctions of subproblems (or subgoals).
AND/OR graphs are an important tool for describing the search spaces generated by many search problems, including those solved by logic-based theorem provers and expert systems~\cite{luger2005artificial}.
In the particular case of planning in production systems, an AND/OR graph is defined as a compact representation of all possible assembly plans of a given semi-finished product \cite{de1990and}.
An AND/OR graph can represent the parallel execution of assembly operations and show the time dependence of operations that can be executed in parallel.
It is ideally suited for parallelized multi-agent assembly.
Moreover, the graph can take non-determinism and uncertainty into consideration via the availability of various branches leading to the desired solution \cite{hawkins2014anticipating}, \cite{russell2016artificial}, \cite{luger2005artificial}.
In this paper, an AND/OR graph as described in the FlexHRC framework~\cite{darwish2017flexible} is adopted to design a flexible human-robot collaboration architecture in an industrial scenario. 
The AND/OR graph aids in achieving the objective \ref{obj1} described in Section \ref{Sec:introduction} to allow robots to reactively adapt to human operator actions. 

\begin{figure}
    \centering
    \includegraphics[scale = 0.4]{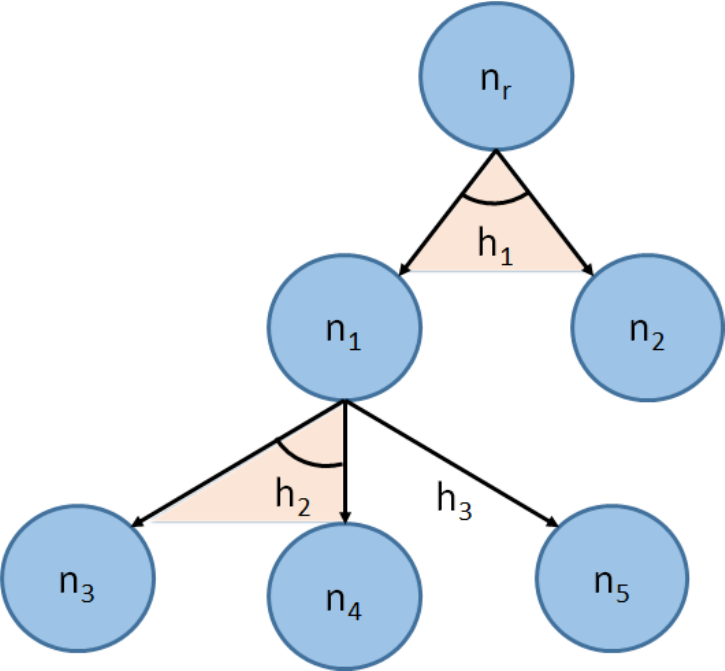}
    \caption{A generic AND/OR graph with 6 nodes and 3 hyper-arcs, where $h_1$ and $h_2$ are \textit{AND} hyper-arcs while $h_3$ is \textit{OR} hyper-arc.}
    \label{fig:andor}
\end{figure}

More formally, an AND/OR graph $G(N, H)$ is a data structure where $N \in \{n_1, n_2, \dots n_{|N|}\}$ represents the set of \textit{nodes}, and $H \in \{h_1, h_2, \dots h_{|H|}\}$ signifies the set of \textit{hyper-arcs}. 
On a semantic level, nodes represent a unique state related to the cooperation task, whereas an hyper-arc represents the possible actions performed by the agents to reach a particular state, i.e., the transitional relationships between the states connected by that specific hyper-arc.
The complete set of definitions and symbols related to the AND/OR graph formalism is presented in Table \ref{tab:and-or-table}.
Nodes follow the classical definitions of graph-like structures:
the \textit{root node} is the \textit{singleton} node at the top of the graph, which represents the final goal of the cooperation process,
\textit{child nodes} are nodes directly connected to another node when moving away from the root, whereas \textit{parent nodes} adhere to the converse notion of the child node.
A \textit{leaf} node is a node that has no child nodes.
A hyper-arc $h \in H$ induces the state transition between the set of child nodes $N_c(h)$ and the singleton set made up of one parent node $N_p(h)$ such that:
\begin{equation}
    h\ :\ N_c(h) \rightarrow N_p(h).
    \label{eq:childparent}
\end{equation}
Since the root node represents the goal state, the cooperation process can be mapped to a graph traversal procedure from a subset of leaf nodes to the root node \textit{optimally}.
A complete discussion about the theory of AND/OR graphs is out of the scope of this paper, and the reader is advised to the treatment in~\cite{luger2005artificial}.
Nodes and hyper-arcs are associated with weights defining the associated costs, namely $(w_{n_1}, w_{n_2}, \dots, w_{n_{|N|}})$ and $(w_{h_1}, w_{h_2}, \dots, w_{h_{|H|}})$ for nodes and hyper-arcs, respectively.
Figure \ref{fig:andor} portrays a generic AND/OR graph with 6 nodes termed $n_r, n_1, n_2, \dots, n_5$ and three hyper-arcs $h_1, h_2, h_3$.
Node $n_r$ is the root node, while $n_2, n_3, n_4, n_5$ are leaf nodes.
Hyper-arc $h_1$ establishes an \textit{AND} relationship between nodes $n_1$, $n_2$ and $n_r$, i.e., in order to reach $n_r$ it is necessary to have reached both $n_1$ and $n_2$. 
It can be expressed as $h_1: n_1 \wedge n_2 \rightarrow n_r$, where the $\wedge$ operator defines the \textit{AND} relationship.
A similar relationship holds for hyper-arc $h_2$, which connects $n_1$, $n_3$ and $n_4$ via an \textit{AND} relationship.
In contrast, hyper-arcs $h_2$ and $h_3$ express the \textit{OR} relationship between nodes $n_1$, $n_3$, $n_4$ and $n_5$.
The semantics associated with $h_2$ and $h_3$ is such that in order to reach node $n_1$, \textit{either} node $n_5$ or the couple $n_3$ and $n_4$ (via $h_2$) need to be reached first, i.e., $h_2 \vee h_3 : (n_3 \wedge n_4) \vee n_5 \rightarrow n_1$, where $\vee$ operator signifies the \textit{OR} relationship.
Depending on the associated weights, the graph traversal algorithm chooses the optimal path to reach $n_r$.

\begin{table*}[]
\centering
\caption{Symbols and definitions related to AND/OR graphs.}
\label{tab:and-or-table}
\begin{tabular*}{\textwidth}{ll}
\hline
\textbf{Symbol}			& \textbf{Definition}																				\\ \hline
$n$					& A node in the AND/OR graph																		\\
$N$					& The set of all nodes in the AND/OR graph															\\
$h$					& A hyper-arc in the AND/OR graph																	\\
$H$					& The set of all hyper-arcs in the AND/OR graph														\\
$G(N,H)$				& An AND/OR graph composed of $N$ nodes and $H$ hyper-arcs											\\
$N_c(h), N_p(h)$		& The hyper-arc $h$ connects the set of child nodes $N_c(h)$ to the singleton set of parent nodes $N_p(h)$			\\
$w_{n}$				& Weight/cost of node $n$																		\\
$w_{h}$				& Weight/cost of hyper-arc $h$																		\\
$a$					& An action associated with one or more hyper-arcs in $G$												\\
$A$					& The set of actions associated with with hyper-arc $h$													\\
$e(a)$				& The action $a$ is \textit{finished} by an agent															\\
$d(h)$				& The hyper-arc $h$ is \textit{done} when all actions in $A$ are executed										\\
$s(n)$				& The node $n$ is solved if the state $n$ has been reached				\\
$s(G)$				& An AND/OR graph is solved if $s(n_r)$ holds true														\\
$f(n)$				& The node $n$ is \textit{feasible} iff for at least one solved hyper-arc $h \in H$ exists such that $p(h) = n$ holds true			\\
$f(h)$				& The hyper-arc $h$ is \textit{feasible} if all the child nodes of $h$ are solved and $\neg e(h)$ holds									\\
$H_f$				& The set of all feasible hyper-arcs													\\
$N_f$				& The set of all feasible nodes							\\
$cp$					& A cooperation path in G																			\\
$c(cp)$					& The cost of cooperation path $cp$													\\
$M(cp)$				& The cooperation model associated with $cp$, i.e., the ordered sequence of actions corresponding to $cp$									
\end{tabular*}
\end{table*}

Each hyper-arc $h$ models a set of actions $A$, where an action $a \in A$ can be performed either by a human operator, a robot, or jointly during the cooperation process.
Actions in the \textit{Task Representation} module are defined at a human-centric semantic level.
In this paper, an important assumption is that action allocation is done offline based on the capabilities of each agent.
For instance, an action called \texttt{inspect}, which is the visual and physical inspection of the part, is always assigned to the human operator, while another action \texttt{grasp} is assigned to the robot.
If an action needs more than one responsible agent, such an action is called \textit{joint action}, an example being \texttt{handover}, which involves the transfer of the part from a \textit{giver} agent to a \textit{receiver} agent \cite{CARFI2019}. 
If the order in which to execute actions in $A$ is important, $A$ is defined as an ordered set such
that: 
\begin{equation}
A = (a_1, \dots , a_{|A|}\ ; \preceq),    
\end{equation}
that is a \textit{temporal sequence} is assumed in the form $a_1 \preceq a_2 \preceq \dots \preceq a_{|A|}$, where $\preceq$ is a precedence operator.
Initially, all the actions in $A$ are labelled as \textit{unfinished}, i.e., $\neg e(a)$ for all $a \in A$.
When an action $a$ is executed successfully, we label it as \textit{finished}, i.e., $e(a)$.
If all actions in $A$ are finished, then the corresponding $h$ is done, which is denoted by $d(h)$.
In addition, if an ordering is induced, $d(h)$ holds \textit{iff} the temporal execution sequence is satisfied, i.e., $\forall a \in A$, if $e(a)$ then $d(h)$ holds true. 

Nodes can be either \textit{solved} or \textit{unsolved}.
A node $n \in N$ is \textit{solved} when the state represented by $n$ has been reached, therefore denoted by $s(n)$. The necessary condition for $s(n)$ is at least one hyper-arc $h \in H$ to this node, i.e., $p(h) = n$, and $h$ is done;
otherwise, the node is denoted as \textit{unsolved}, i.e., $\neg s(n)$.
It is noteworthy that \textit{leaf nodes} in AND/OR graphs are initialized as solved or unsolved at the beginning of the cooperation process, depending on the initial state of the cooperation.

Using these definitions, we introduce the notion of \textit{feasibility} of nodes and hyper-arcs.
During graph traversal, any node $n \in N$ is denoted as \textit{feasible}, which is indicated as $f(n)$, \textit{iff} there is at least one solved hyper-arc $h \in H$ to it such that $p(h) = n$;
otherwise, $n$ is \textit{unfeasible}, which is denoted as $\neg f(n)$.
Similarly, a hyper-arc $h$ is \textit{feasible}, i.e., $f(h)$, \textit{iff} for each node $n \in N_c(h)$, then $n$ is solved and $\neg e(h)$ holds.
Once a hyper-arc $h_i \in H$ is solved, all other feasible hyper-arcs $h_j \in H \setminus \{h_i\}$, which share with $h_i$ at least one child, i.e., $N_c(h_i) \cap N_c(h_j) \neq \emptyset$, are marked unfeasible.
This is done so as to prevent the cooperation process to consider irrelevant alternatives repeatedly, thus speeding up the online search.

In order to reach the objective of traversing the graph to the root node, an optimal path in terms of overall cost must be chosen.
As mentioned above, each node $n$ and hyper-arc $h$ is associated with a cost, namely $w_{n}$ and $w_{h}$, respectively.
These costs or weights are assigned depending on a number of parameters such as the difficulty of the associated actions, human operator preferences, or time-to-completion.
A \textit{cooperation path} $cp$ in $G$ can be defined as a unique way to connect leaf nodes to the root node, such that:
\begin{equation}
    cp = (n_1, \dots, n_k, h_1, \dots, h_l).
\end{equation}

According to the structure of a particular AND/OR graph, there may be multiple cooperation paths or, in other words, multiple ways of solving the problem.
Each cooperation path $cp$ is associated with a traversal cost $c(cp)$, which defines how effortful following the path is, on the basis of the involved node and hyper-arc weights, such that:
\begin{equation}
    c(cp) = \sum_{j=1}^k w_{n_j} + \sum_{j=1}^l w_{h_j}. 
    \label{costfunction}
\end{equation}
The different cooperation paths are ranked according to their associated overall costs.
Two cooperation paths are said to be \textit{equal} \textit{iff} they share the same nodes and hyper-arcs, whereas they are \textit{equivalent} \textit{iff} they share the same overall cost.
Depending on the optimal cooperation path, i.e., the one minimizing the overall cost depending on node and hyper-arc weights, the robot may start moving or waiting for human actions. 

During graph traversal, the sets of feasible nodes $N_f$ and hyper-arcs $H_f$ are defined.
If at any stage $N_f \cup H_f = \emptyset$ holds, then the cooperation task is stopped and considered as \textit{failed} because there are no further feasible nodes or hyper-arcs leading to the root node.
However, when the root node $n_r$ is solved, i.e., $s(n_r)$ holds, then $G$ is labelled as solved, i.e., $s(G)$ holds and the cooperation task is complete.
Unlike other search techniques, it must be noted that online AND/OR graphs do not require \textit{full observability} of all the states.
Full knowledge of the robot workspace and human actions is not necessary.
In fact, only the knowledge of the feasible nodes and hyper-arcs is required for making the cooperation task progress.

The AND/OR graph traversal algorithm is organised in two phases, the first offline and the second online.
The offline phase loads the description of the AND/OR graph defining a cooperation process and initializes all nodes, hyper-arcs and paths.
In the online phase, when the AND/OR graph is queried with the last set $N_s$ of solved nodes and the set $H_s$ of solved hyper-arcs, the algorithm updates the status of all nodes and hyper-arcs, as well as the cost of the paths, and provides the sets $N_f$ and $H_f$ of currently feasible nodes and hyper-arcs.
The offline and online phases have been described in~\cite{darwish2017flexible}.

The \textit{Task Manager} invokes a \textit{Sequential Planner}, i.e., the plan is a totally ordered sequence of actions and the optimal plan is the shortest one, i.e., the one with the lowest cost \cite{sebastia2001stella}. 
However, it must be noted that various other options would be equally legitimate for the planner.
The planner receives the set of feasible states and state transitions from the \textit{Task Representation} module, it determines the sequence of actions in each state and the corresponding state transitions, it grounds the semantic action parameters, and assigns the actions to robots or human operators such that the objective of cooperation is maintained.
The \textit{Task Manager} module also reactively adapts to varying operator actions online.
In the \textit{Task Manager}, we define the planning problem to find the ordered sequence of actions $(a_1,\dots, a_n)$ from the initial state to the goal such that the state transition is executed~\cite{darwish2017flexible}, as noted above.
However, in this paper, we do not discuss how such a sequence can be obtained, and we consider it to be provided \textit{a priori}, otherwise assumed to be already computed.

Given the set of feasible states or state transitions and the associated costs, the \textit{Task Manager} attempts to achieve an optimal path to the goal by either proactively deciding the state transition to follow (i.e., \textit{proactive decision making}), or by following human preferences (i.e., \textit{reactive adaptation}).
Proactive decision making is the process of selecting the state with the minimum cost, i.e., the optimal state, according to \eqref{costfunction}, grounding the literals, assigning the actions to the human operator or the robot, and the examination of the optimal state execution.
Reactive adaptation works on the principle of choosing states based on the present scenario without any effect of past experiences.
The \textit{optimal path} is the one with the least cost as defined by \eqref{costfunction} to reach the cooperation goal.
In each state, the \textit{Task Manager} executes the ordered sequence of actions corresponding to the state transition to the next optimal state. 
It allocates the actions to either the human operator or the robot, and transmits the action command to either the \textit{Human Interface} or the \textit{Robot Interface} module, respectively.

During the cooperation process, when an acknowledgement is received from the \textit{Robot Interface} module, the next action in the sequence is evaluated to determine the responsible agent.
If the action must be performed by the robot, then the \textit{Task Manager} sends a message to the \textit{Robot Interface}, which sends the corresponding commands to the \textit{Robot Controller}.
If the action needs to be performed by the human operator, the \textit{Task Manager} sends a message to the \textit{Human Interface} module asking the operator to perform the task.
Note that in this case, the human operator is free to choose whether to follow the instructions from the \textit{Human Interface} module or not.
Hence, when an acknowledgement is received from the \textit{Human Interface} module, a check is done to determine whether the human operator is still following the same cooperation path. 
Thus, during action execution, if operators decide to perform another feasible action, the \textit{Task Manager} reactively adapts to give priority to their decision, and works towards achieving the goal state.
For a complete description on the underlying algorithms involved in the \textit{Task Representation} and the \textit{Task Manager} modules, the reader is advised to look at previous work~\cite{darwish2017flexible},~\cite{darvish2018interleaved}.
\section{Experimental Evaluation}
\label{sec:experiment}

\subsection{Experimental Process}
\begin{figure*}[t!]
\begin{center}
\subfloat[]{
    \resizebox*{9cm}{5cm}{\includegraphics{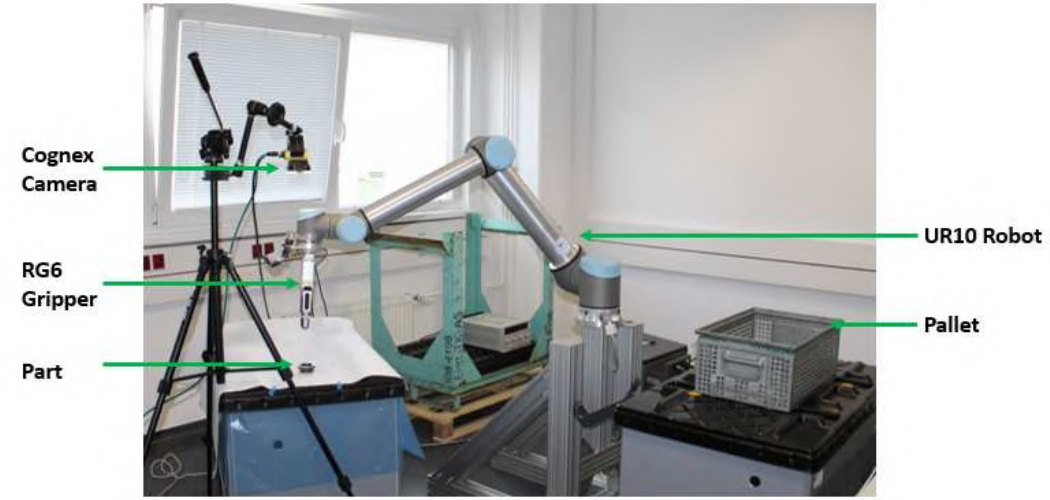}}
    \label{fig:hardwaresetup}
}
\subfloat[]{
    \resizebox*{8cm}{5cm}{\includegraphics{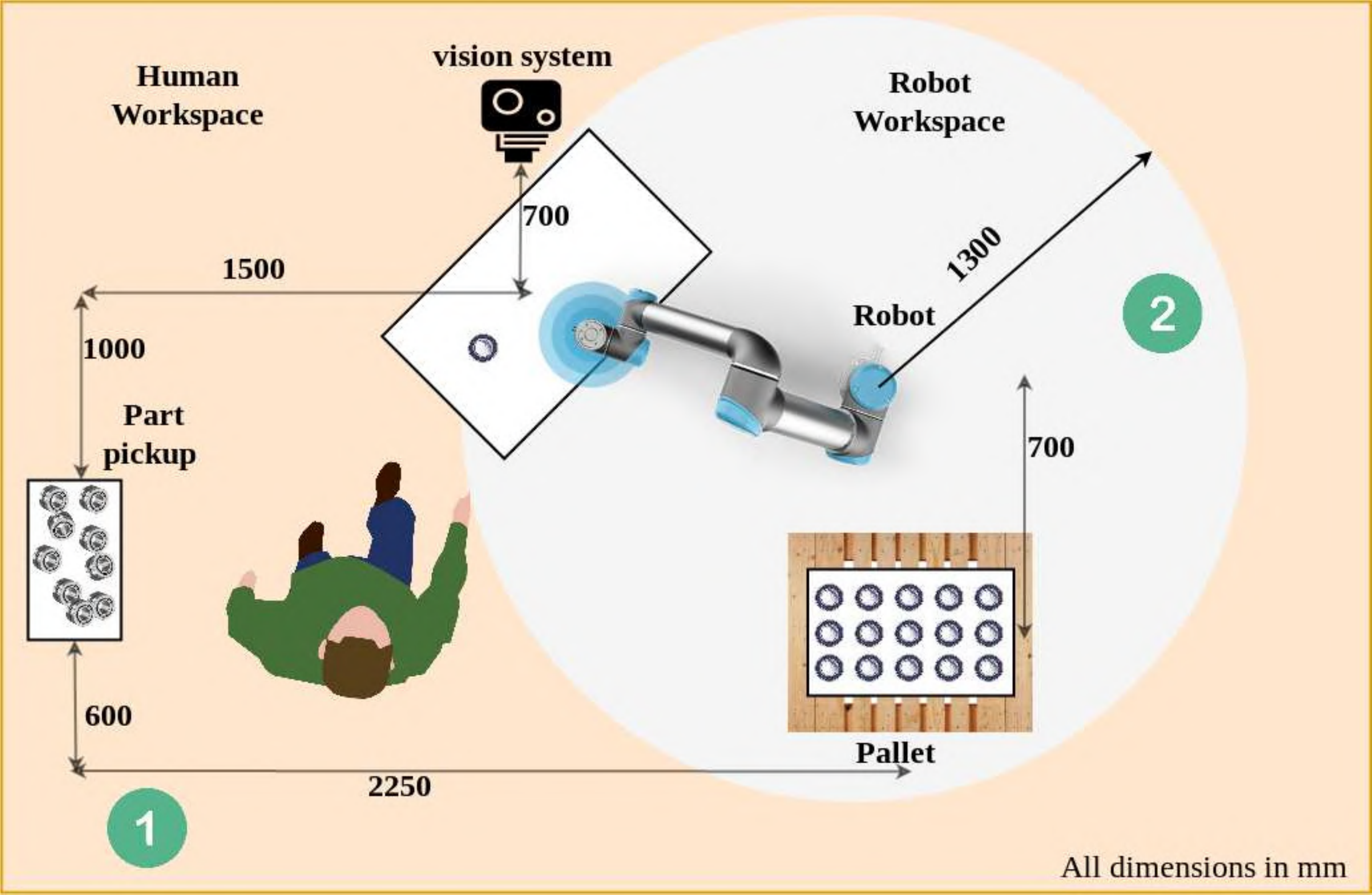}}
    \label{fig:tp-view}
}
\caption{(a) Hardware setup for the experiments showing the UR10 robot, a pallet, a part, the RG6 gripper, and the Cognex vision system (the part pick-up box is not shown). (b) Schematic top view of the experiment setup to show the workspace limits of the robot and the human operator. The area marked $1$ in orange is the shared area, whereas the area marked $2$ in white is the collaboration area.}
\label{hardware-setup}
\end{center}
\end{figure*}

\begin{figure}[t!]
    \centering
    \includegraphics[width= 6cm]{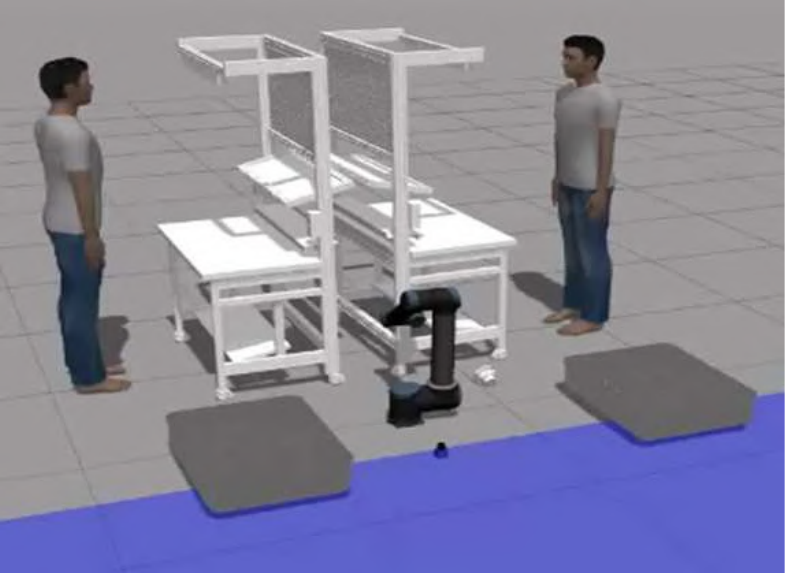}
    \caption{The simulated environment in Gazebo showing the robot, the workstation table, a pallet, and human operators.}
    \label{fig:digTwin}
\end{figure}

We have evaluated the proposed framework using a 6 degrees-of-freedom (DoF) UR10 manipulator from Universal Robot, which is equipped with a RG6 OnRobot Gripper, a Cognex Vision system for determining object poses, and a workstation using a ROS Indigo based middleware on an Ubuntu 14.04 OS running on a 64-bit Intel i7 processor with 8GB RAM.
The source code 
\footnote{\url{https://github.com/EmaroLab/industrialRobot_task_planning}}
developed for the \textit{representation level} adopts C++, whereas in the \textit{interface} and \textit{perception levels} is developed in Python.
The \textit{Robot Controller} module is an adapter that modifies the ROS Industrial \texttt{ur-modern-driver}~\cite{andersen2015optimizing} with additional ROS services to account for gripper control and inverse kinematics for grasping the part. 
In particular, since the \texttt{ur-modern-driver} provides joint level control for the UR10 robot, we perform inverse kinematics calculations using \texttt{URScript}~\cite{URmanual} commands to the robot's internal controller.
In order to enforce rapid prototyping for solutions based on our system, we developed a Gazebo-based, ROS-compatible simulation \cite{koenig2004design}.
The simulated scenario is presented in Figure~\ref{fig:digTwin}.
The auxiliary fixed structures such as the workstation table and the pallet in Figure~\ref{fig:digTwin} are 3D models provided by Schaeffler, whereas robot models are extracted from the ROS Industrial repository. 
The hardware setup is shown in Figure \ref{hardware-setup}(a), whereas the top view is shown in Figure \ref{hardware-setup}(b) in order to better clarify the robot and the human operator workspaces.
The experimental setup resembles the target application described in Section \ref{Sec:introduction}, and shown in Figure \ref{fig:application_process}. 
It is noteworthy that a human operator can enter the robot workspace and touch the robot during its motion.
The robot is mounted on a fixed structure.

\begin{figure*}[t!]
\begin{center}
\subfloat[]{
    \resizebox*{6cm}{!}{\includegraphics{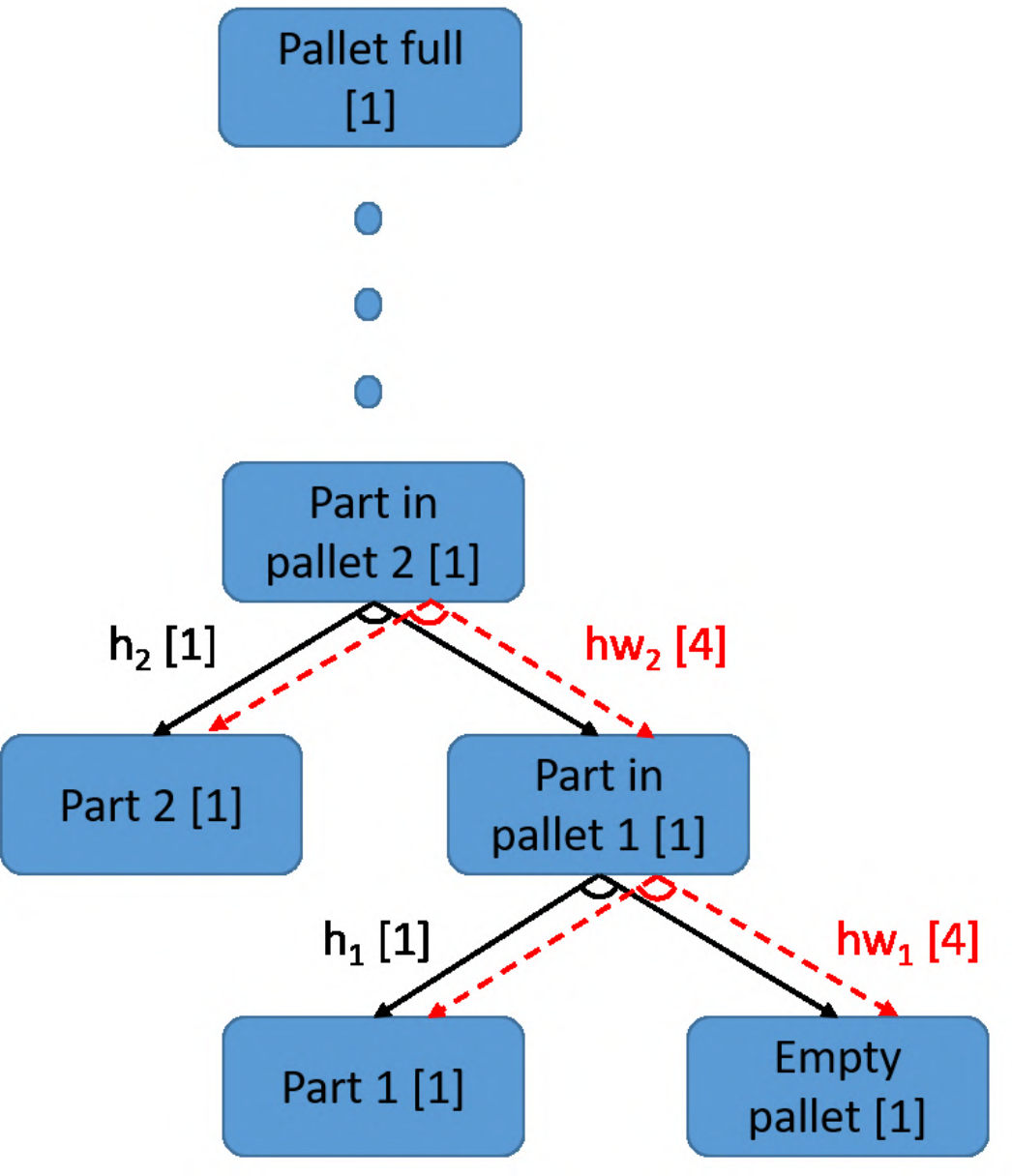}}
}
\subfloat[]{
    \resizebox*{10cm}{!}{\includegraphics{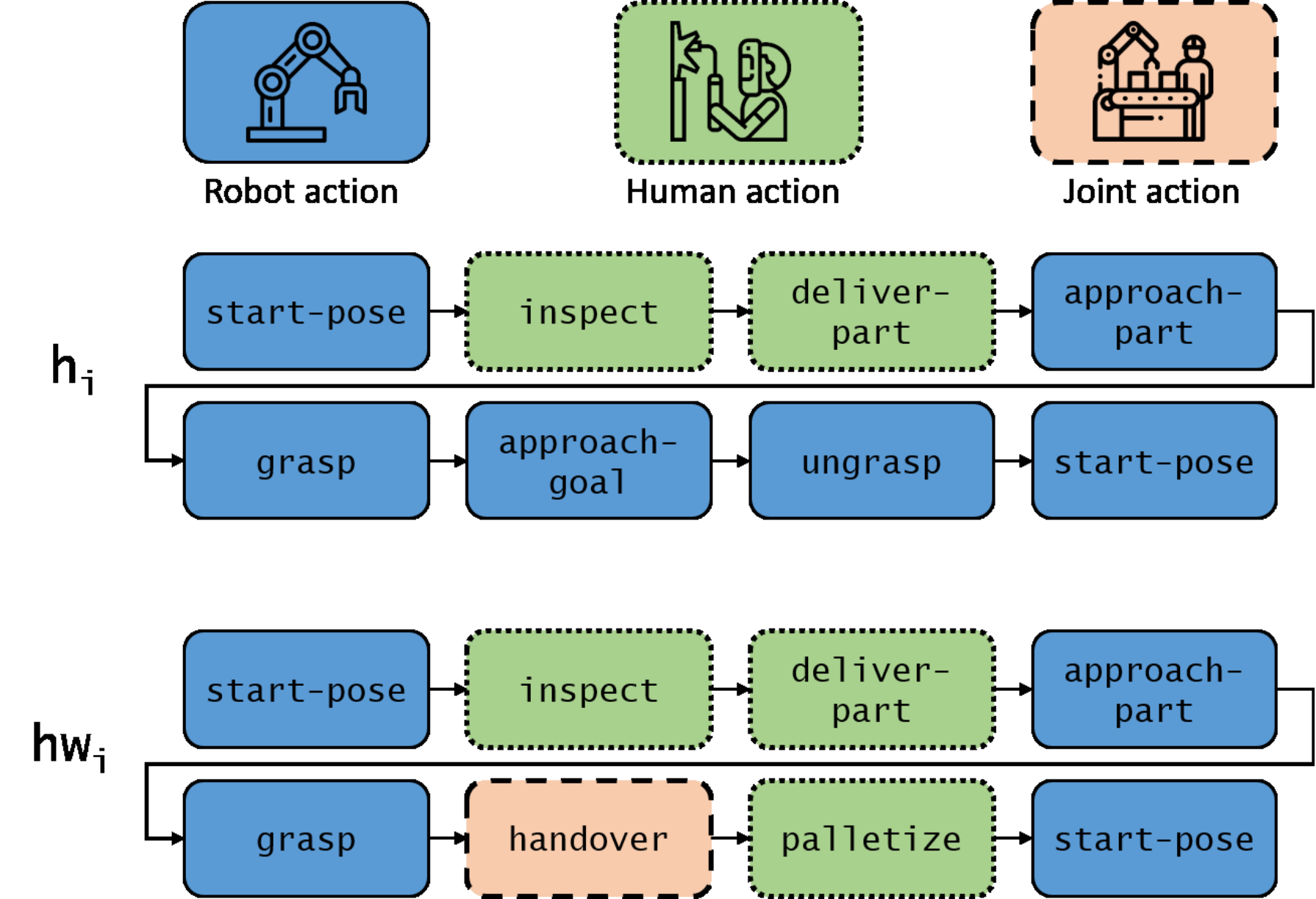}}
}
\caption{(a) The AND/OR graph to traverse from the initial configuration (called \texttt{empty-pallet}, when a \texttt{part} is available) to the final state (called \texttt{pallet-full}). The set of hyper-arcs $h_i$ and $hw_i$ are represented by the solid-black arrows and the dashed-red arrows, respectively. The weights of each node and hyper-arc are shown in brackets. The graph is truncated to avoid unnecessary repetitions. (b) A sequence of actions corresponding to each hyper-arc of type $h$ and $hw$, and the agent associated with each action.}
\label{fig:overall-andor}
\end{center}
\end{figure*}

The AND/OR graph corresponding to the complete cooperation task is shown in Figure \ref{fig:overall-andor}(a).
As anticipated in the previous Section, we consider the root node $n_r$ as the goal state, which in this case corresponds to the completion of the palletization process consisting of 15 parts, which is referred to as \texttt{pallet-full} state.
At each step, there are two possible cooperation paths denoted by hyper-arcs $h_i$ and $hw_i$ where $i \in (1, 2, \dots, 15)$.
Hence there are $2^{15}$, i.e., 32786 possible paths to reach the goal state.
A human operator can perform the following actions: 
\begin{itemize}
    \item \texttt{inspect} is the careful tactile and visual inspection of the part;
    \item \texttt{deliver-part} is the transferring of the part to the robot by the human operator;
    \item \texttt{palletize} is the process of stacking the part inside the pallet.
\end{itemize}
The robot can perform the following actions:
\begin{itemize}
    \item \texttt{start-pose} requires the robot to move its end-effector to a previously defined start pose; 
    \item \texttt{approach-part} assumes that the robot finds the most suitable grasping points for the part, and moves the end-effector in a pose to grasp it;
    \item \texttt{grasp} is executed when the end-effector has reached the grasping point and the object is graspable;
    \item \texttt{approach-goal} corresponds to robot motion actions approaching the goal pose to place the part, and eventually release it;
    \item \texttt{ungrasp} is executed to release the part when it has been properly positioned.
\end{itemize}
Finally, the \texttt{handover} action must be performed \textit{jointly} by both the human operator and the robot, according to well-established sequential patterns \cite{CARFI2019}.

Typical action sequences are shown in Figure \ref{fig:overall-andor}(b), parametrized with respect to generic hyper-arcs $h_i$ and $hw_i$, for all 15 parts.
A generic hyper-arc $h$ involves a human operator to perform \texttt{inspect} and \texttt{deliver-part} to the robot, whereas the robot is responsible to perform the palletization.
A hyper-arc $hw$ involves the so-called \textit{intervention case} where human operators stop the robot, retrieve the part from the robot gripper, and perform palletization themselves.
In general, we assign a higher cost to hyper-arcs in the form of $hw$, i.e., $w(hw_i) = 4$ for all $i$, in comparison to hyper-arcs in the form of $h$, for which $w(h_i) = 1$.
The reason is two-fold:
on the one hand, a hyper-arc in the form of $hw$ is not required at every cycle and may be performed at random by the human operator;
on the other hand, it helps enforce objective \ref{obj1} (related to ergonomy), i.e., the human operator should not carry heavy parts whenever possible.
As anticipated above, the human operator can obtain information about the cooperation process such as robot and joint actions in a human-friendly format from the \textit{User Interface} module.
The overall task is defined as \textit{failed} in the following situations: 
an action carried out by a human operator is not recognized by the vision system within a time-out period of 2 minutes, a bound that has been heuristically defined;
any of robot actions are unsuccessful; 
the AND/OR graph enters an \textit{unfeasible} state due to communication failures between the perception, the representation and the interface layers.
 
We performed experiments with $10$ employees from the Schaeffler Group, comprising of production line workers and line supervisors.
Examples of such experiments can be found in the accompanying video \footnote{Please refer to: \url{https://youtu.be/mF22PPmwHN0}.}.
It is important to note that this study does not have a pretence of being of any statistical significance, and can be consider as a pilot study.
All the volunteers are males and between age of $29$ to $45$.
Five volunteers had experience working with industrial robots prior to the experiment.
Each \textit{trial} involves $3$ experiments as explained below.
All volunteers performed one trial, hence in total we performed $10$ trials or $30$ experiments.
The experiment is divided into a \textit{familiarization} phase and a \textit{test} phase. 
Prior to the familiarization phase, each volunteer is shown a demo of the experiment by the experiment invigilator.
During the familiarization phase, volunteers are allowed one rehearsal through each experiment supervised and assisted by the invigilator.
During the test phase, volunteers have to perform each experiment without any instructions or assistance from the invigilator.
Their consent for video-taping the experiment during the test phase is taken by the invigilator.
Volunteers are provided with two Likert-scale questionnaires (for pre-experiment and post-experiment), which provide a basis for a subjective evaluation of the system.
Furthermore, data regarding human action time, robot action time, the time taken by \textit{Task Manager}, operator idle time, robot idle time and overall time for execution are also recorded.
These provide objective evaluation criteria.

\begin{figure*}[t!]
\begin{center}
\subfloat[]{
\resizebox*{5cm}{!}{\includegraphics{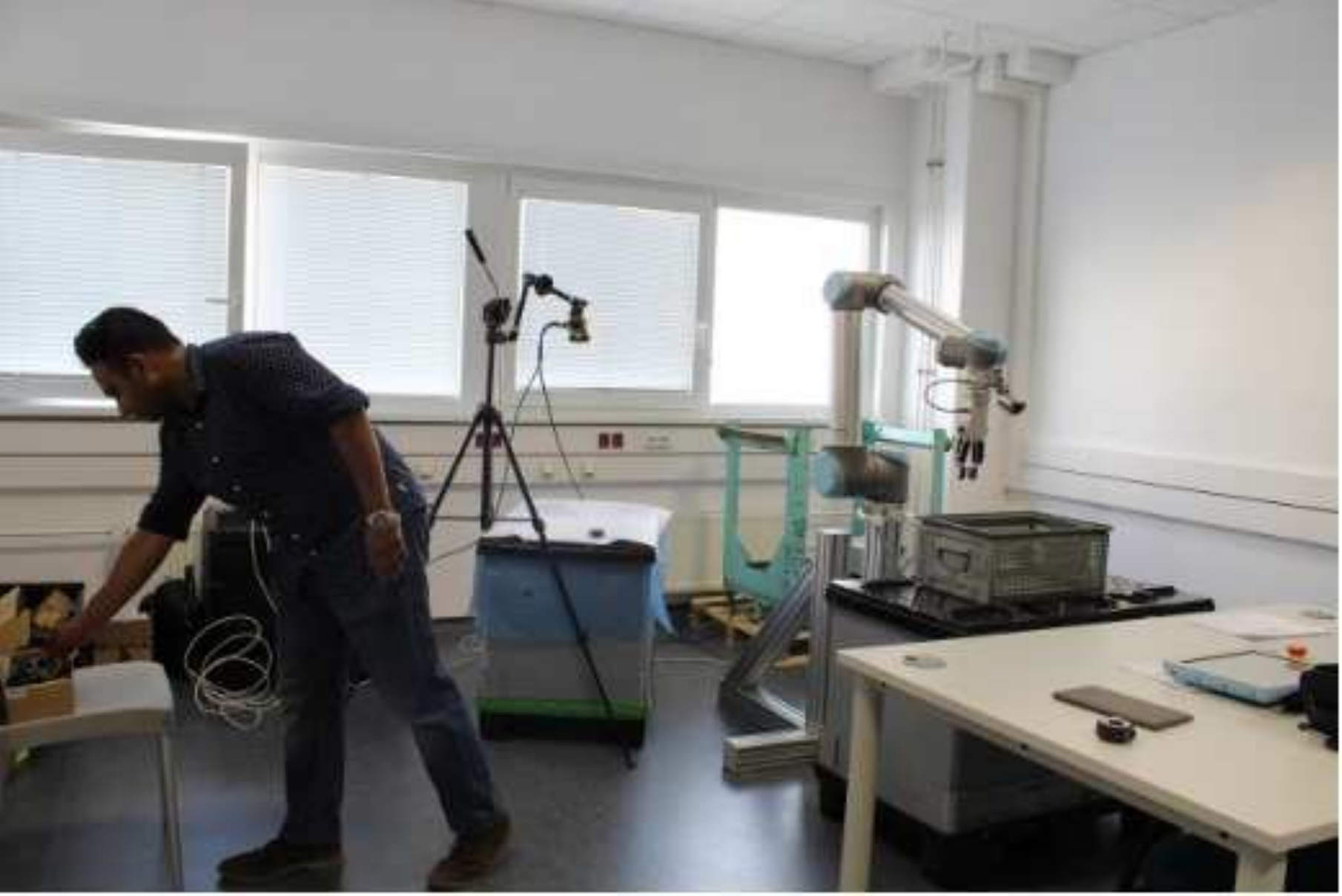}}}
\subfloat[]{
\resizebox*{5cm}{!}{\includegraphics{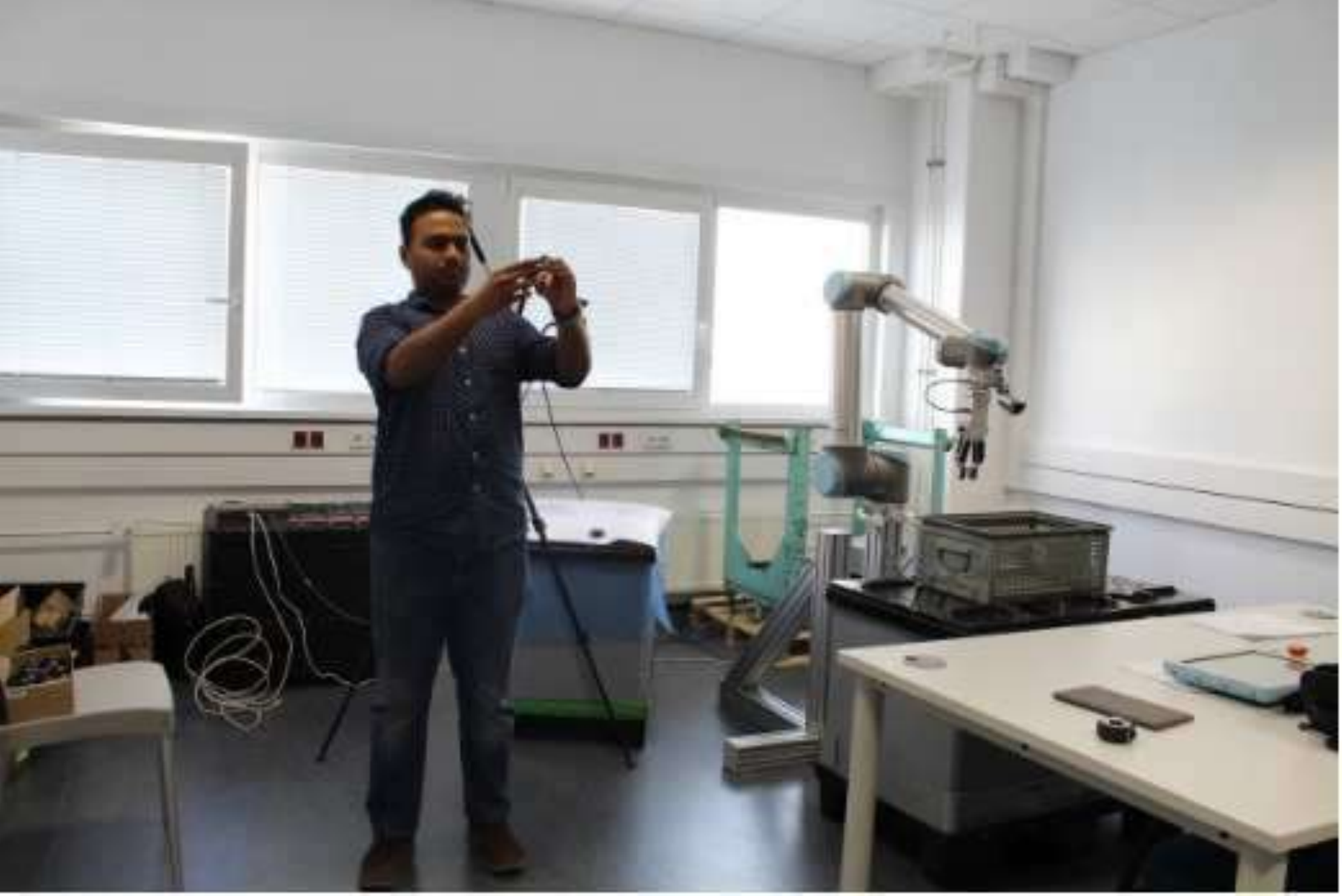}}}
\subfloat[]{
\resizebox*{5cm}{!}{\includegraphics{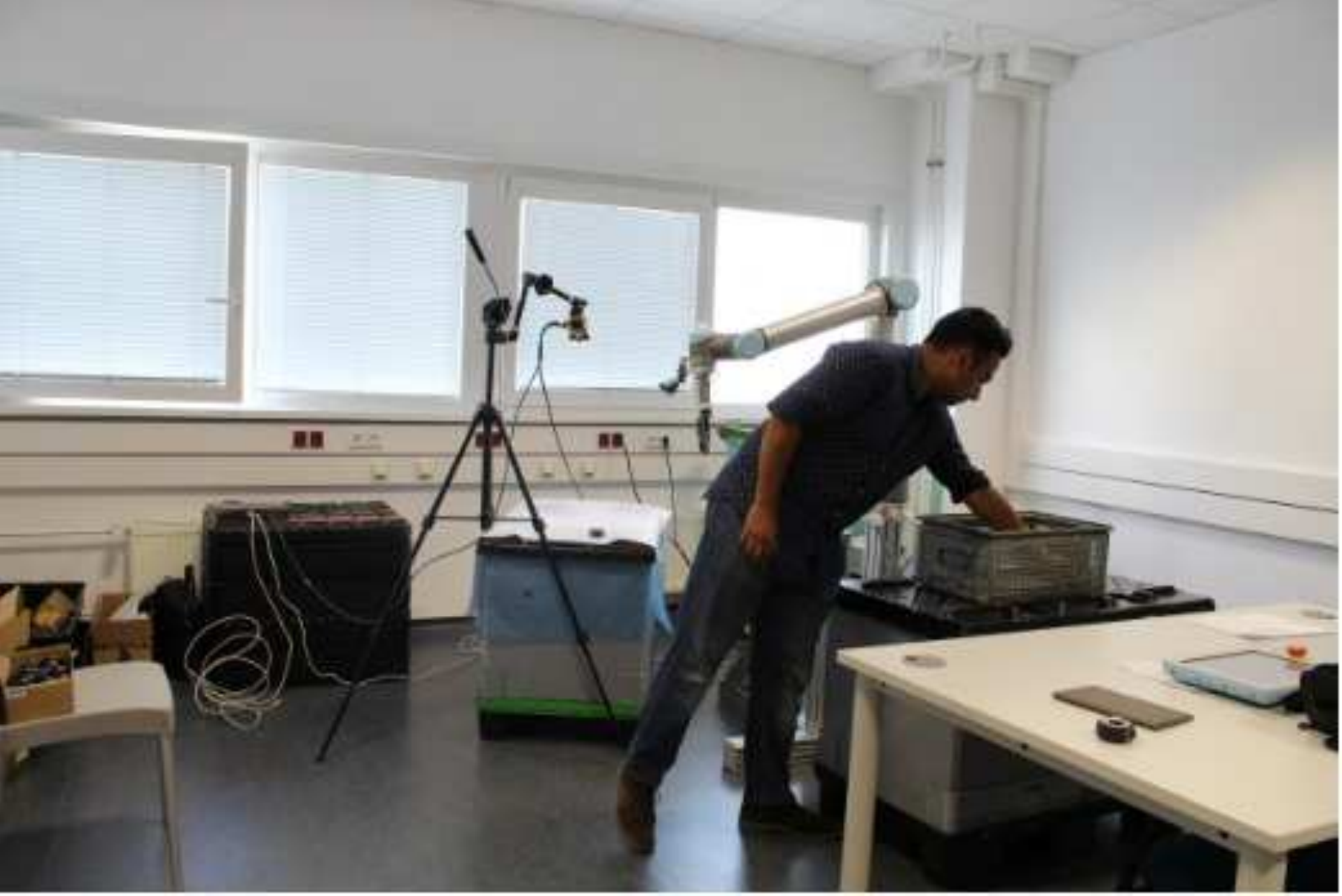}}}
\caption{An action sequence corresponding to the \textit{Manual execution} experiment: (a) the operator receives the part, (b) performs visual and tactile inspection (\texttt{inspect}), (c) performs the palletization of the part (\textit{palletize}). Our system is not operational in this experiment.}
\label{fig:manual_new}
\end{center}
\end{figure*}

\begin{figure*}[t!]
\begin{center}
\subfloat[]{
\resizebox*{4cm}{3cm}{\includegraphics{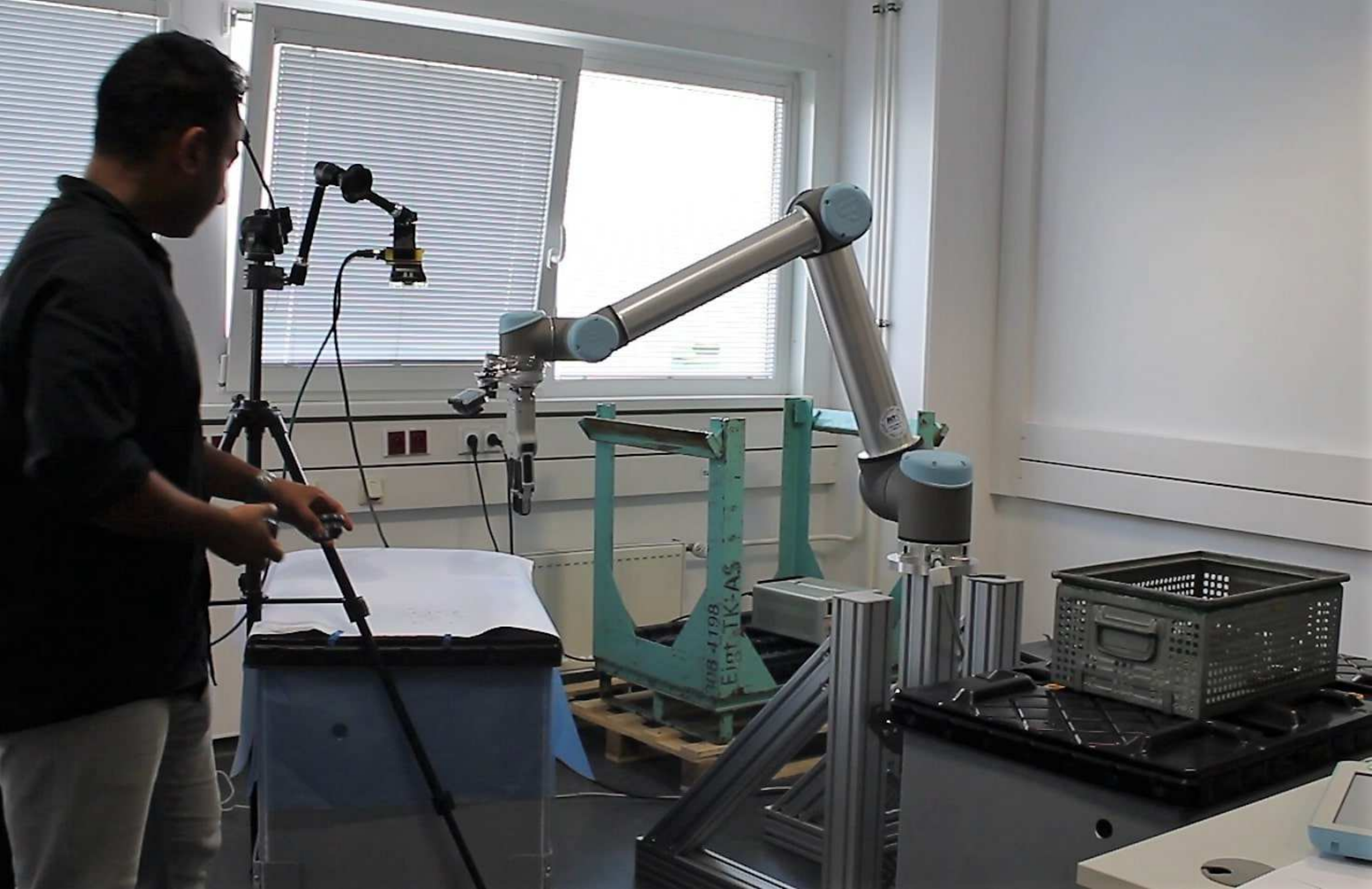}}}
\subfloat[]{
\resizebox*{4cm}{3cm}{\includegraphics{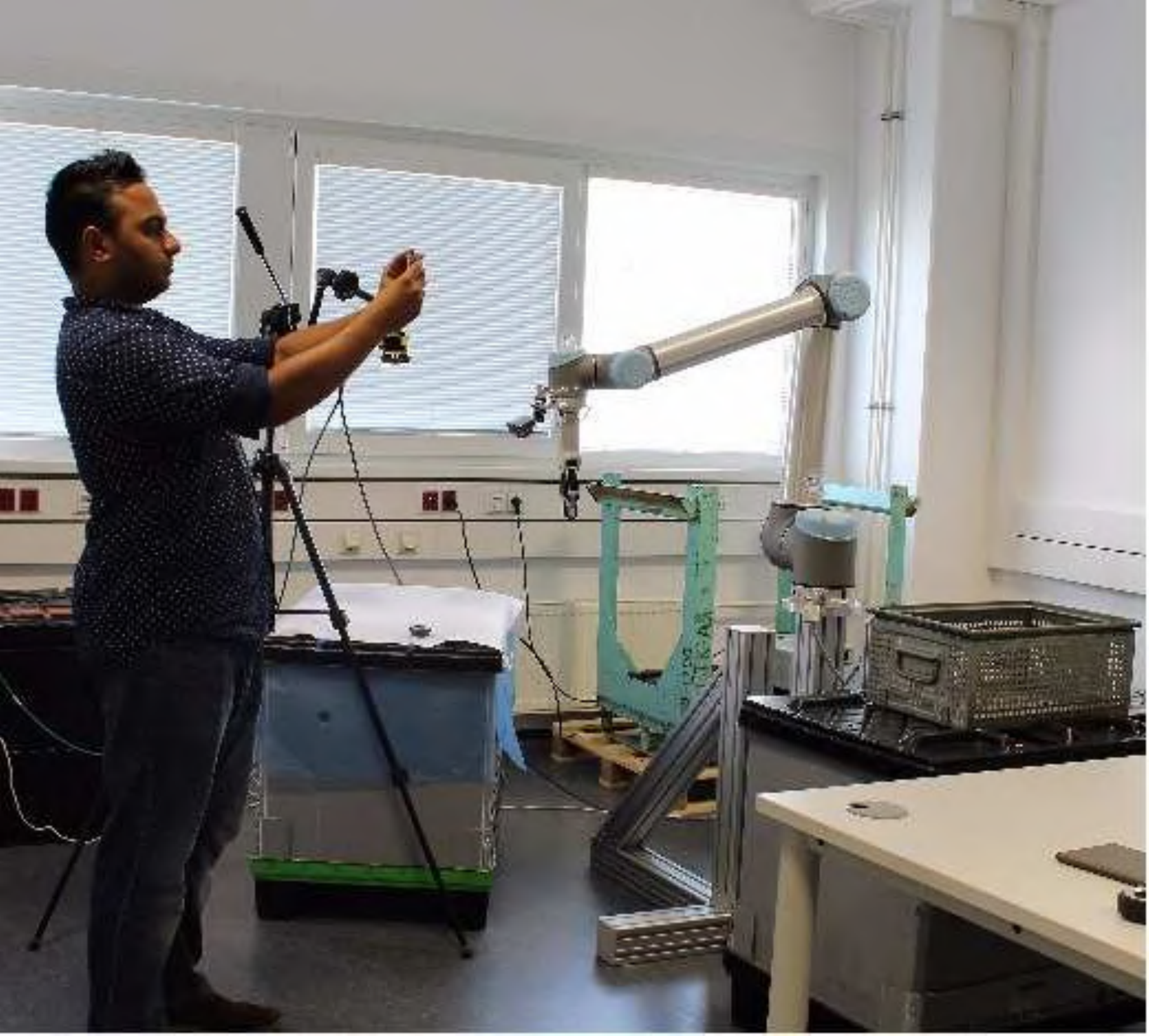}}}
\subfloat[]{
\resizebox*{4cm}{3cm}{\includegraphics{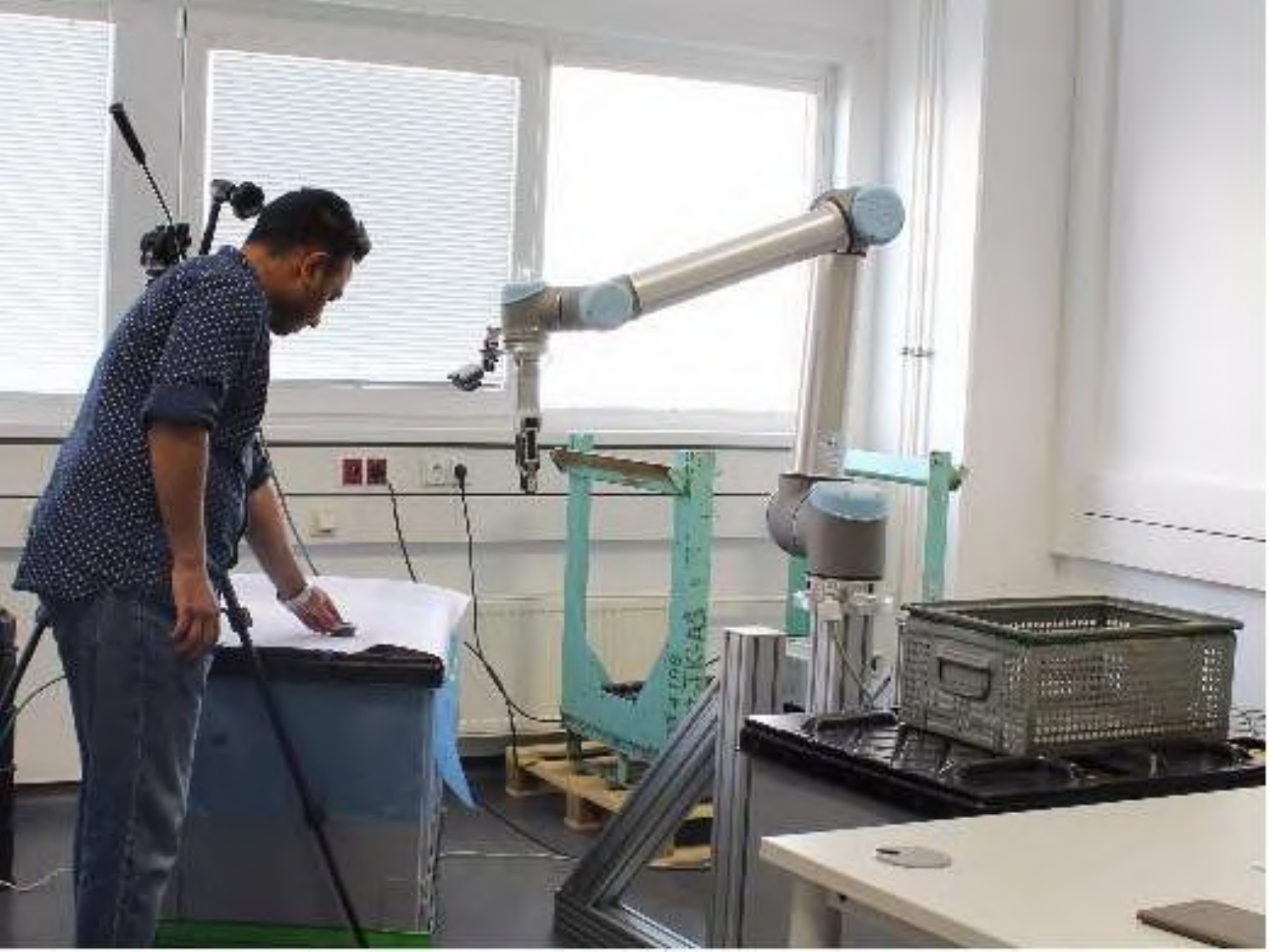}}}
\subfloat[]{
\resizebox*{4cm}{3cm}{\includegraphics{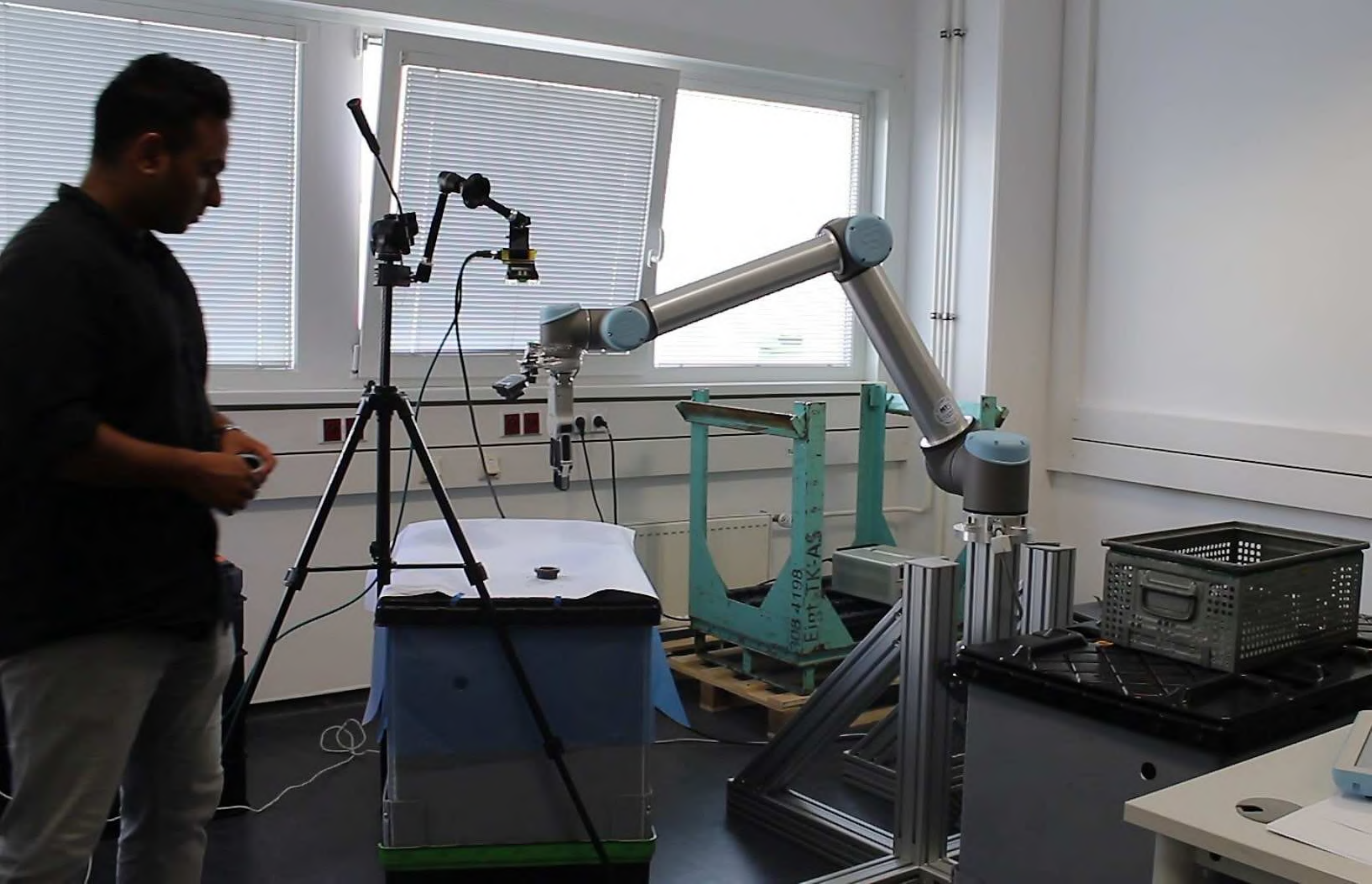}}}\hfill
\subfloat[]{
\resizebox*{4cm}{3cm}{\includegraphics{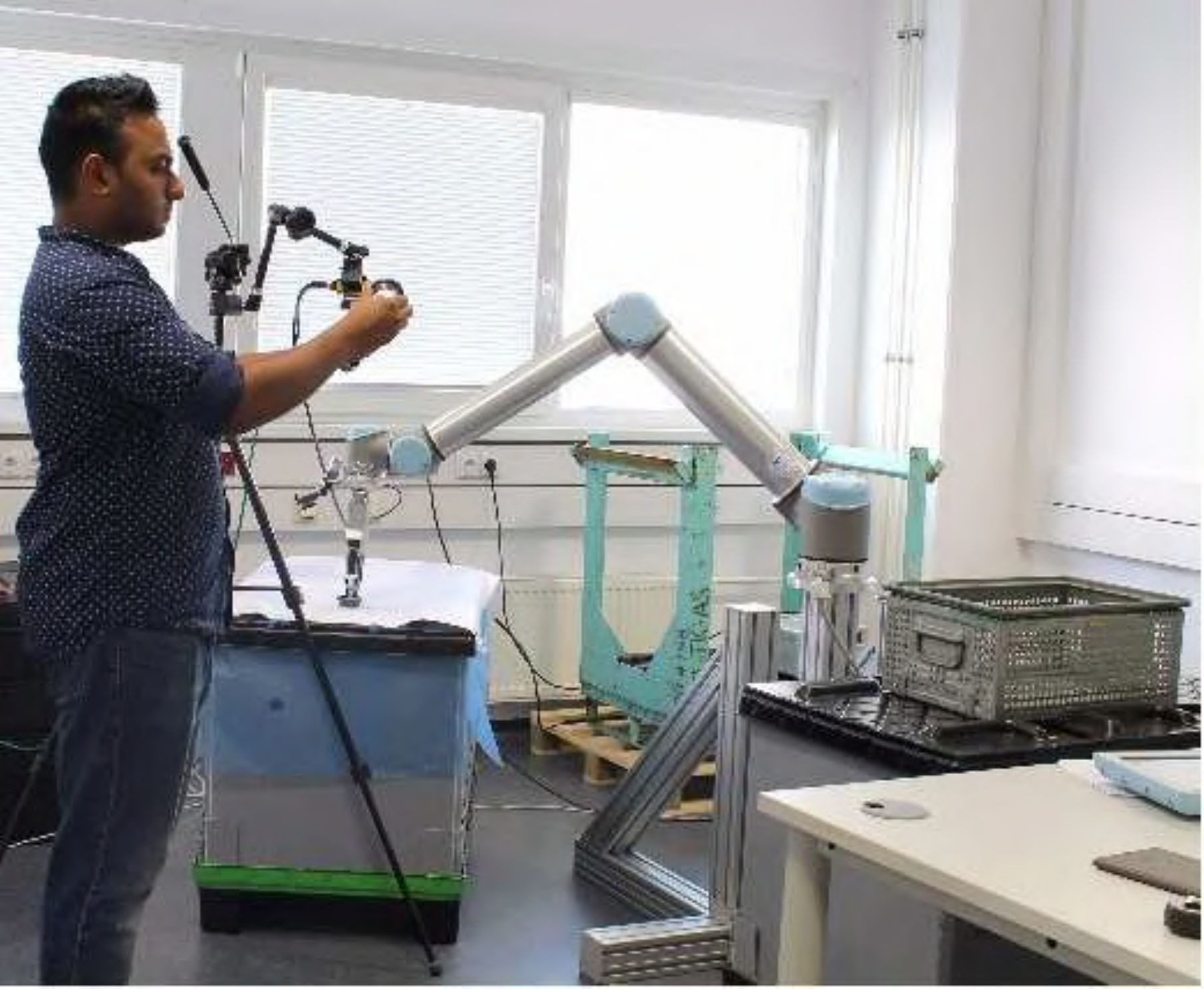}}}
\subfloat[]{
\resizebox*{4cm}{3cm}{\includegraphics{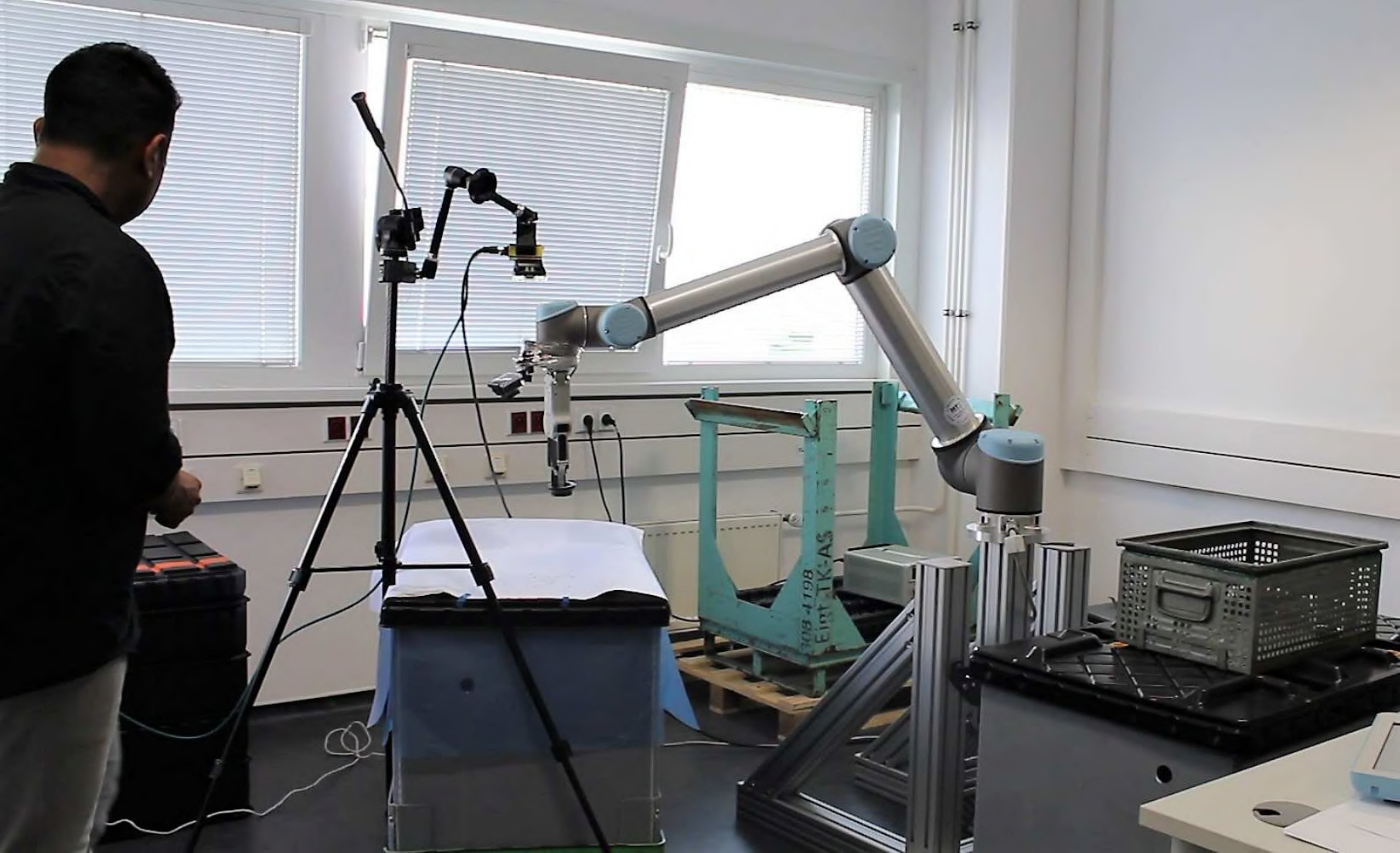}}}
\subfloat[]{
\resizebox*{4cm}{3cm}{\includegraphics{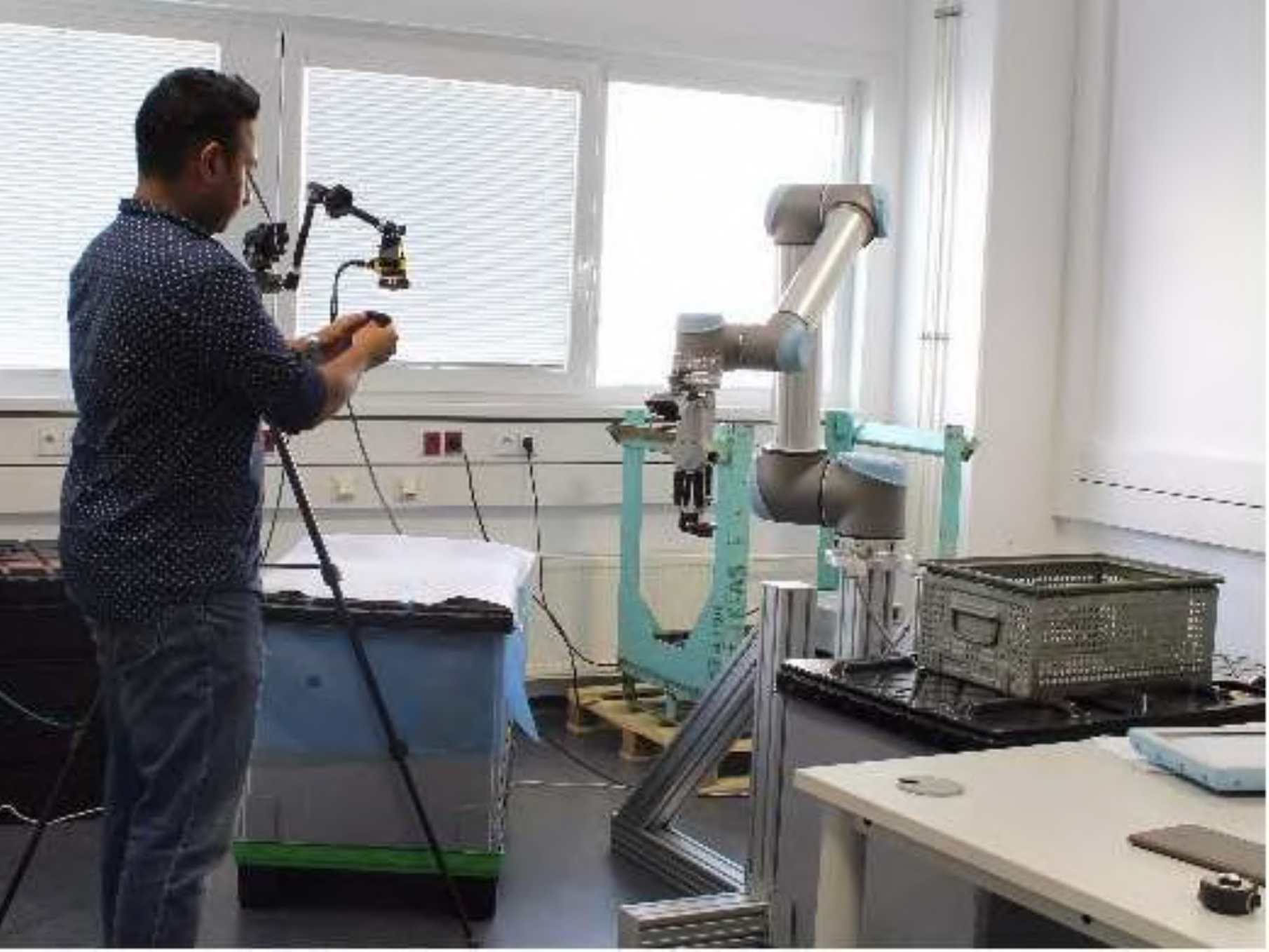}}}
\subfloat[]{
\resizebox*{4cm}{3cm}{\includegraphics{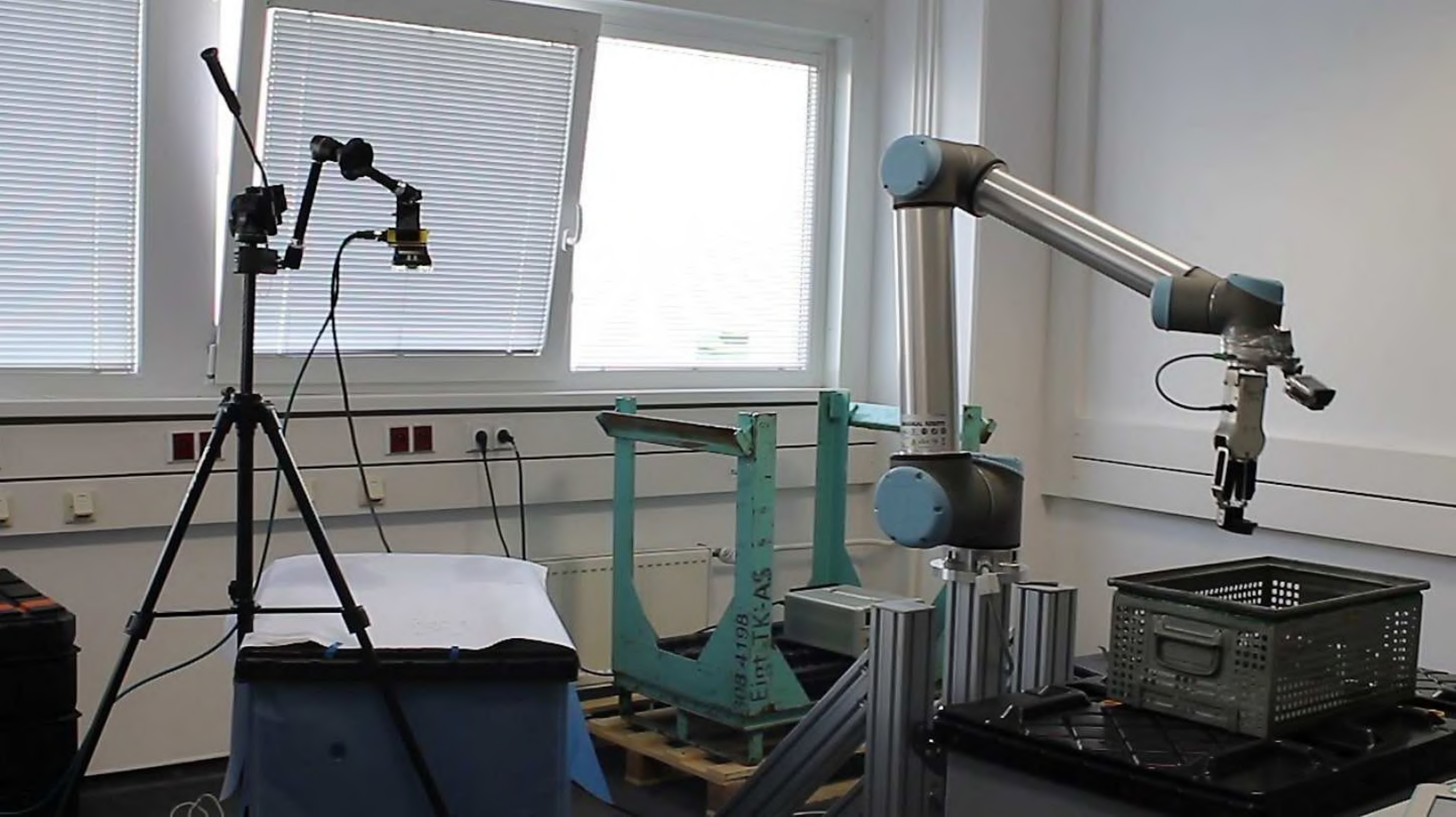}}}\hfill
\subfloat[]{
\resizebox*{4cm}{3cm}{\includegraphics{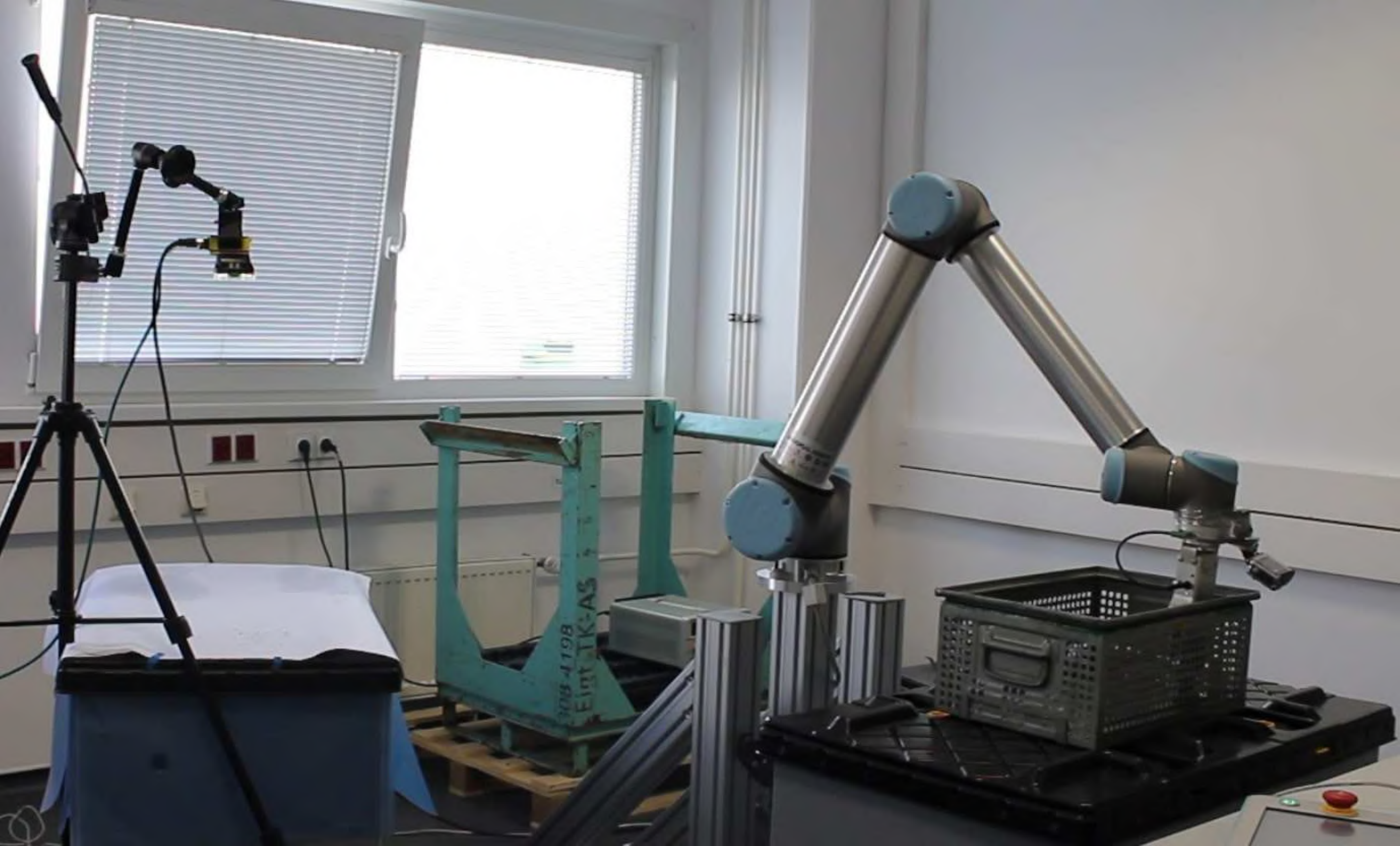}}}
\subfloat[]{
\resizebox*{4cm}{3cm}{\includegraphics{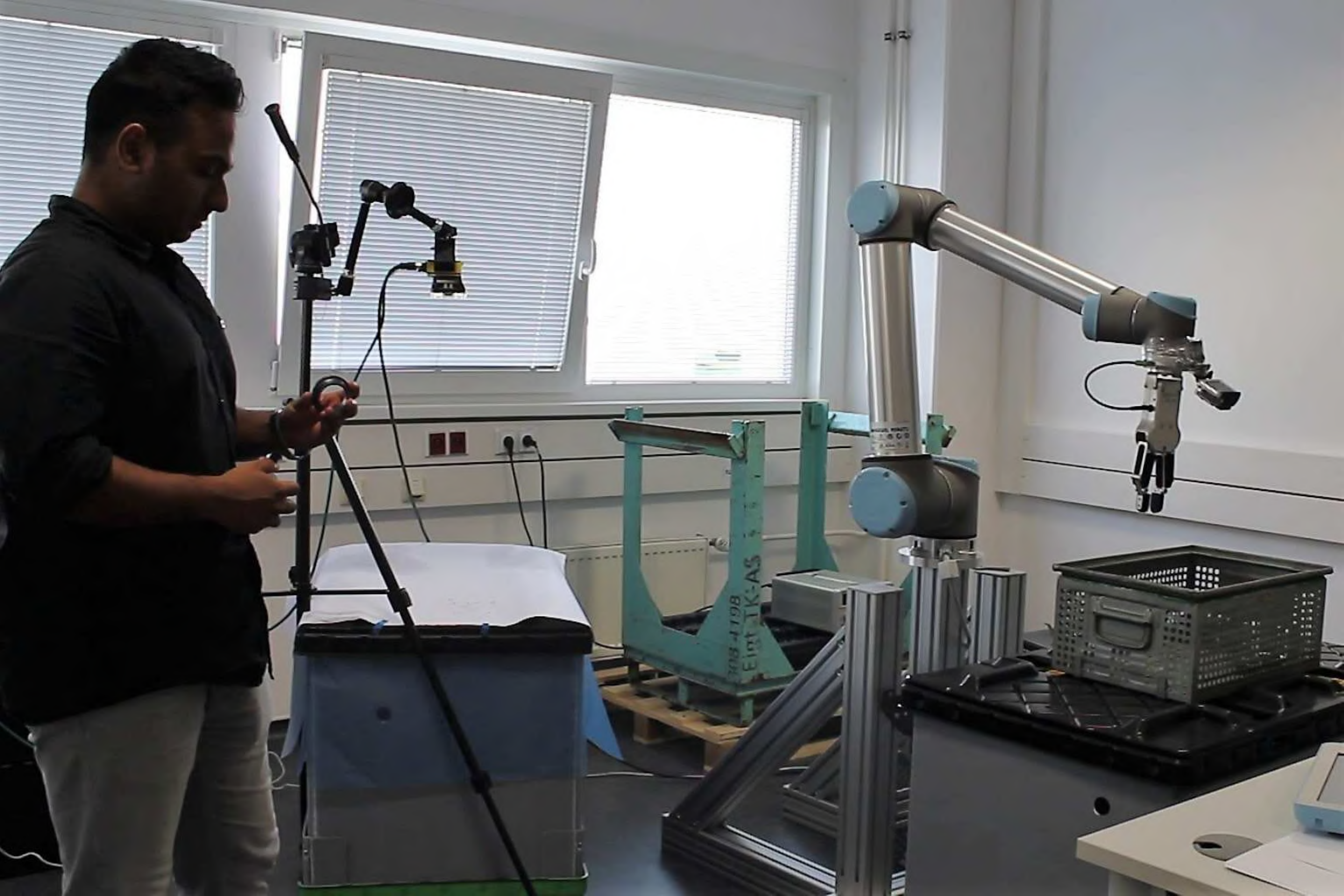}}}
\subfloat[]{
\resizebox*{4cm}{3cm}{\includegraphics{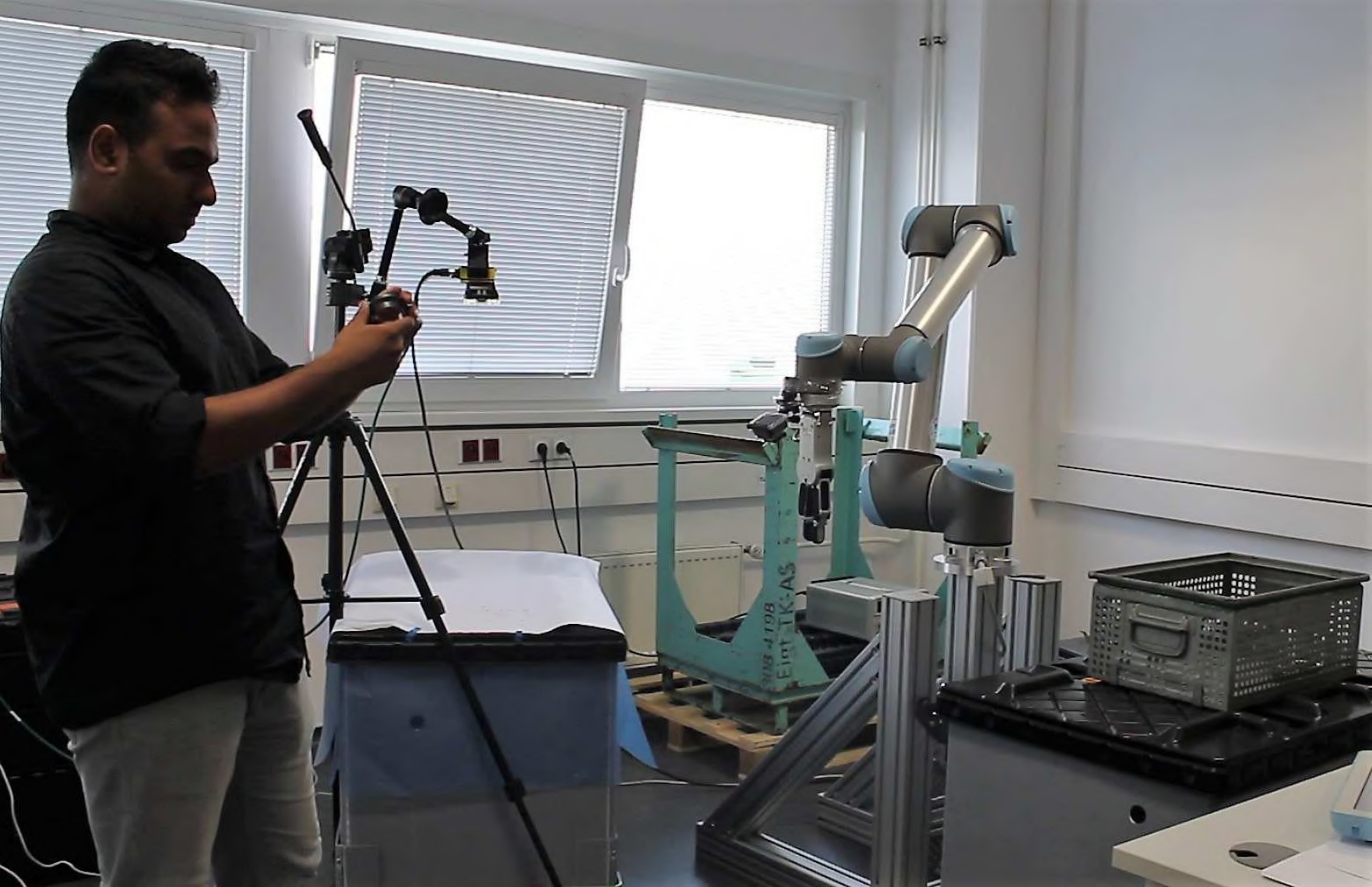}}}
\subfloat[]{
\resizebox*{4cm}{3cm}{\includegraphics{Experiment/handover/startPose.pdf}}}
\caption{An action sequence depicting the \textit{Co-existence} experiment: (a) the robot waits in the start position (\texttt{start-pose}), (b) the human operator performs visual and tactile inspection (\texttt{inspect}), (c) places the part in robot pickup area (\texttt{deliver-part}), (d) the robot approaches the object (\texttt{approach-part}), (e) the robot grasps the object (\texttt{grasp}), (f-h) the robot moves the part to the pallet (\texttt{approach-goal}), (i) the robot places the part inside the pallet box (\texttt{ungrasp}), (i-l) the robot moves back to the start position and process continues (\texttt{start-pose}). These actions correspond to hyper-arcs of type $h$ as described above.}
\label{fig:handover_new}
\end{center}
\end{figure*}

\begin{figure*}[t!]
\begin{center}
\subfloat[]{
\resizebox*{7.5cm}{!}{\includegraphics{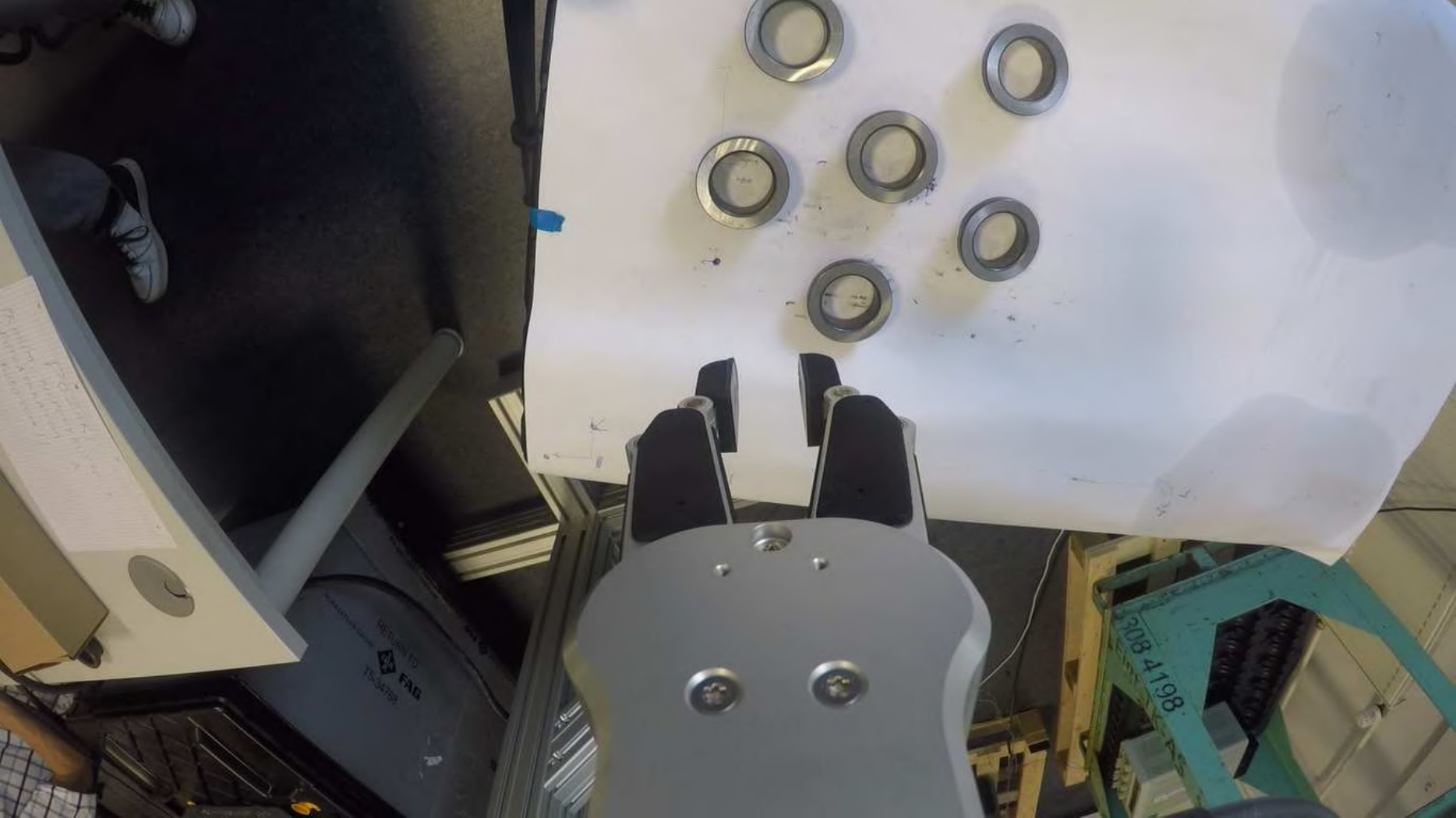}}}
\subfloat[]{
\resizebox*{7.5cm}{!}{\includegraphics{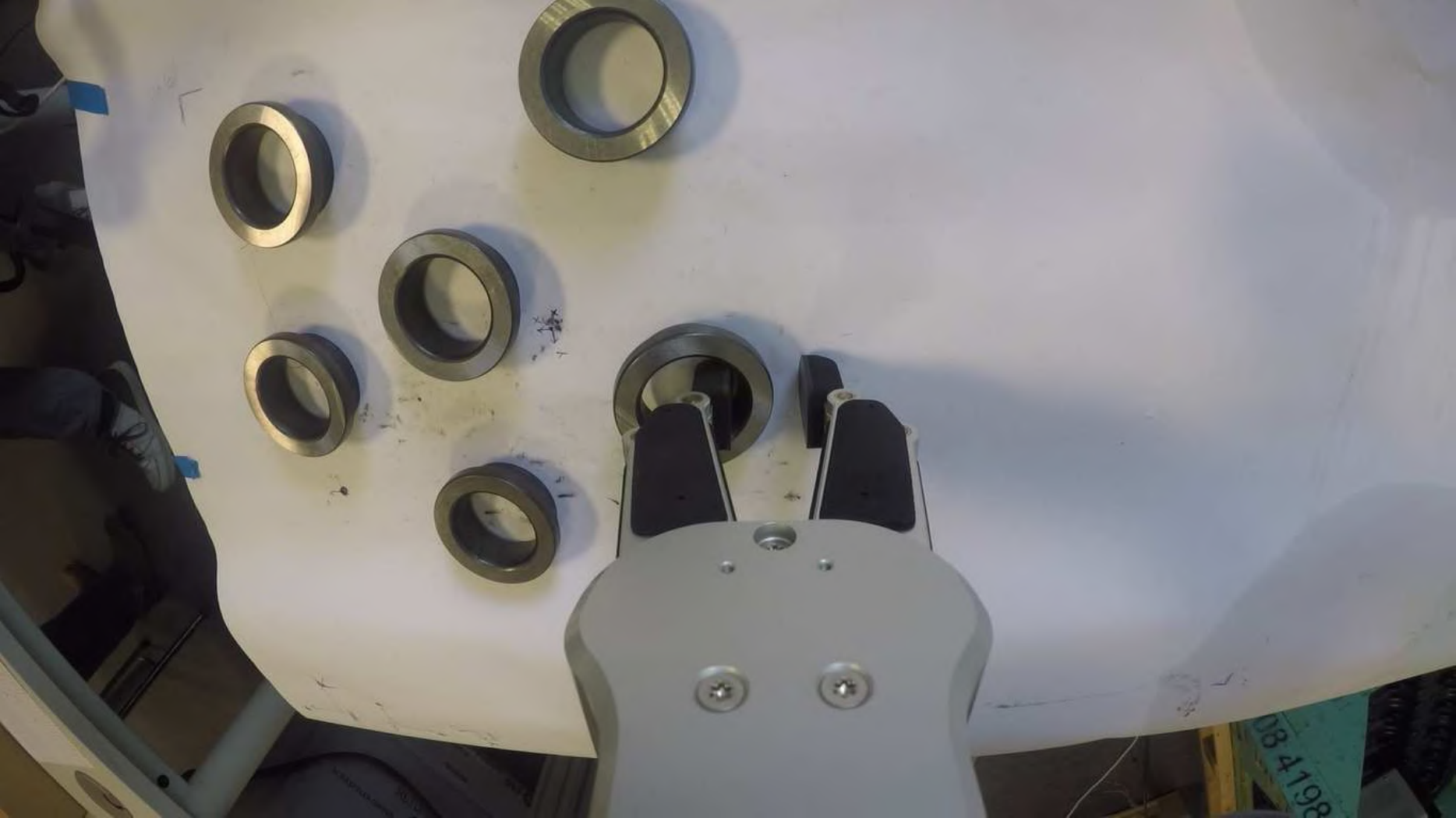}}}
\caption{The system handling multiple parts: (a) the robot waits for the vision module to provide grasp locations from the clutter of parts, (b) the robot moves to grasp the part.}
\label{fig:robust_new}
\end{center}
\end{figure*}

\begin{figure*}[t!]
\begin{center}
\subfloat[]{
\resizebox*{4cm}{3cm}{\includegraphics{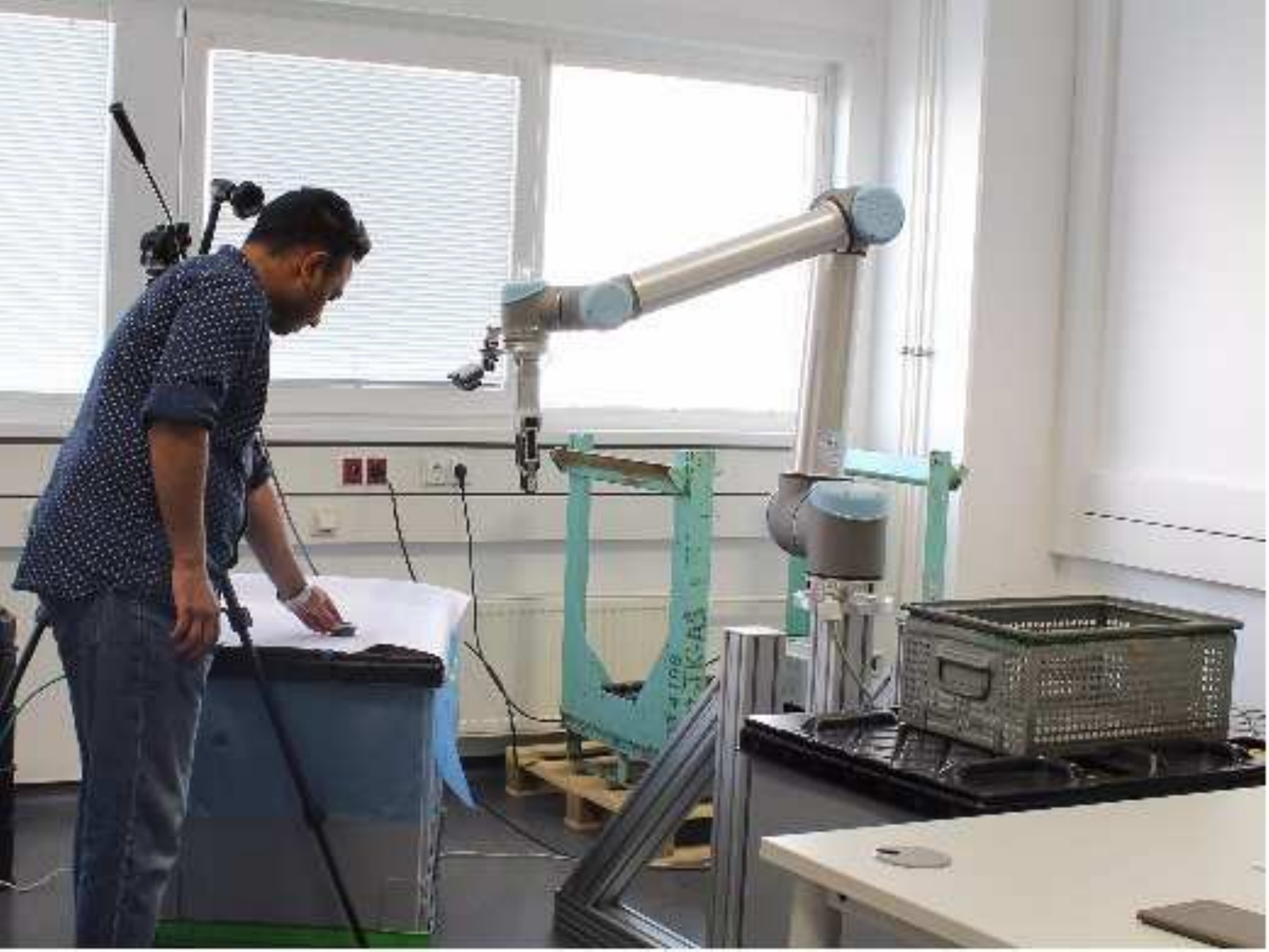}}}
\subfloat[]{
\resizebox*{4cm}{3cm}{\includegraphics{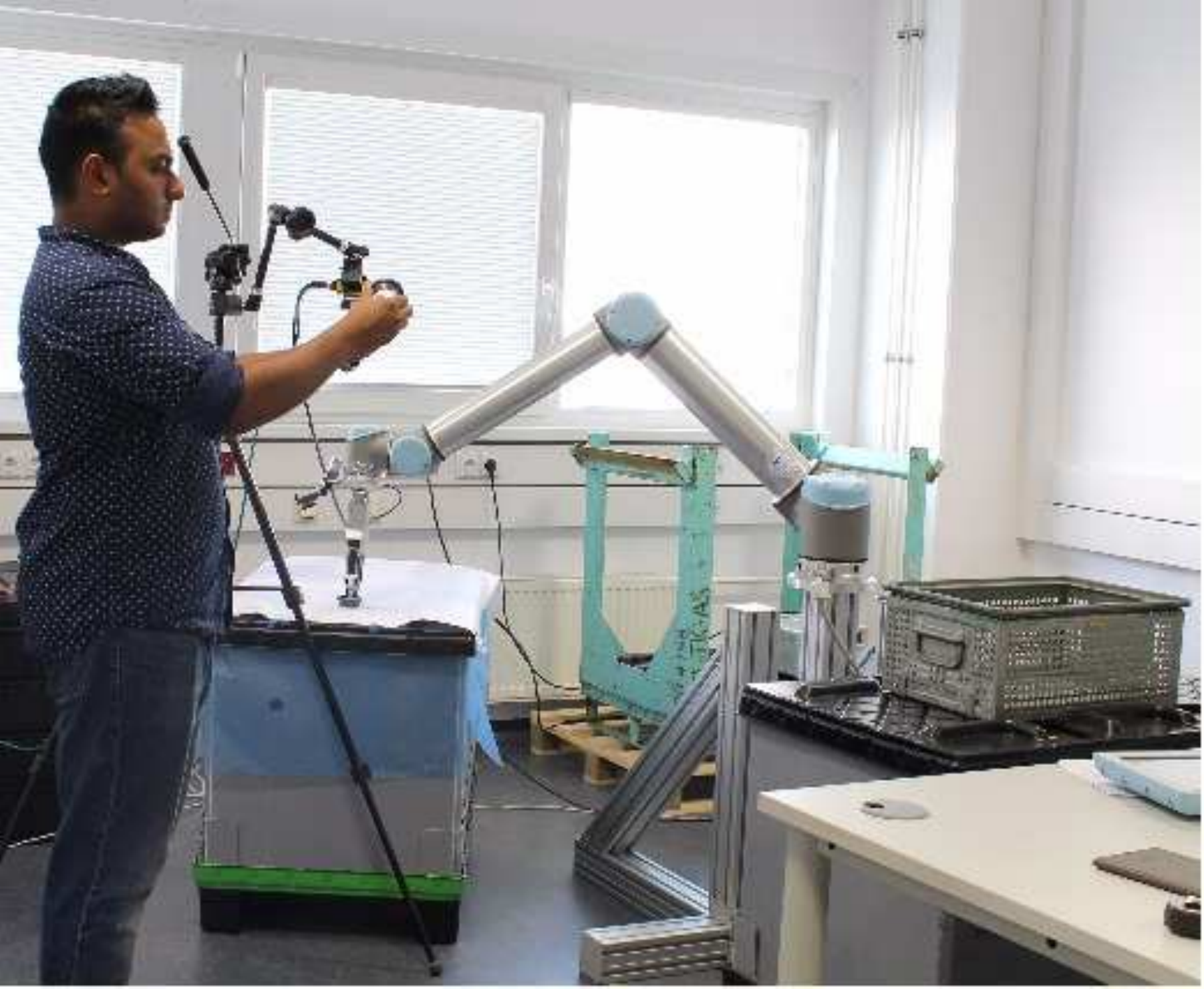}}}
\subfloat[]{
\resizebox*{4cm}{3cm}{\includegraphics{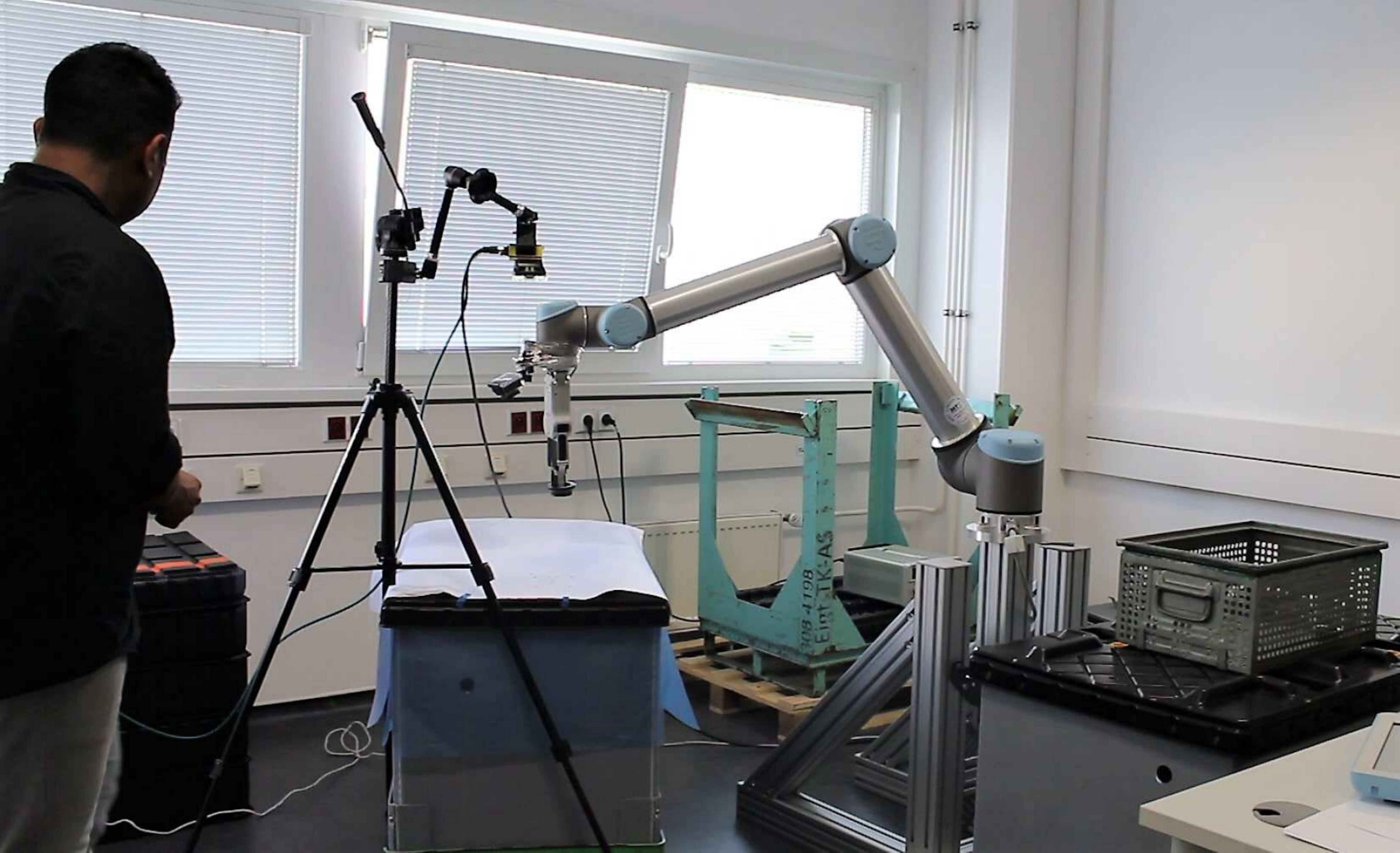}}}
\hfill
\subfloat[]{
\resizebox*{4cm}{3cm}{\includegraphics{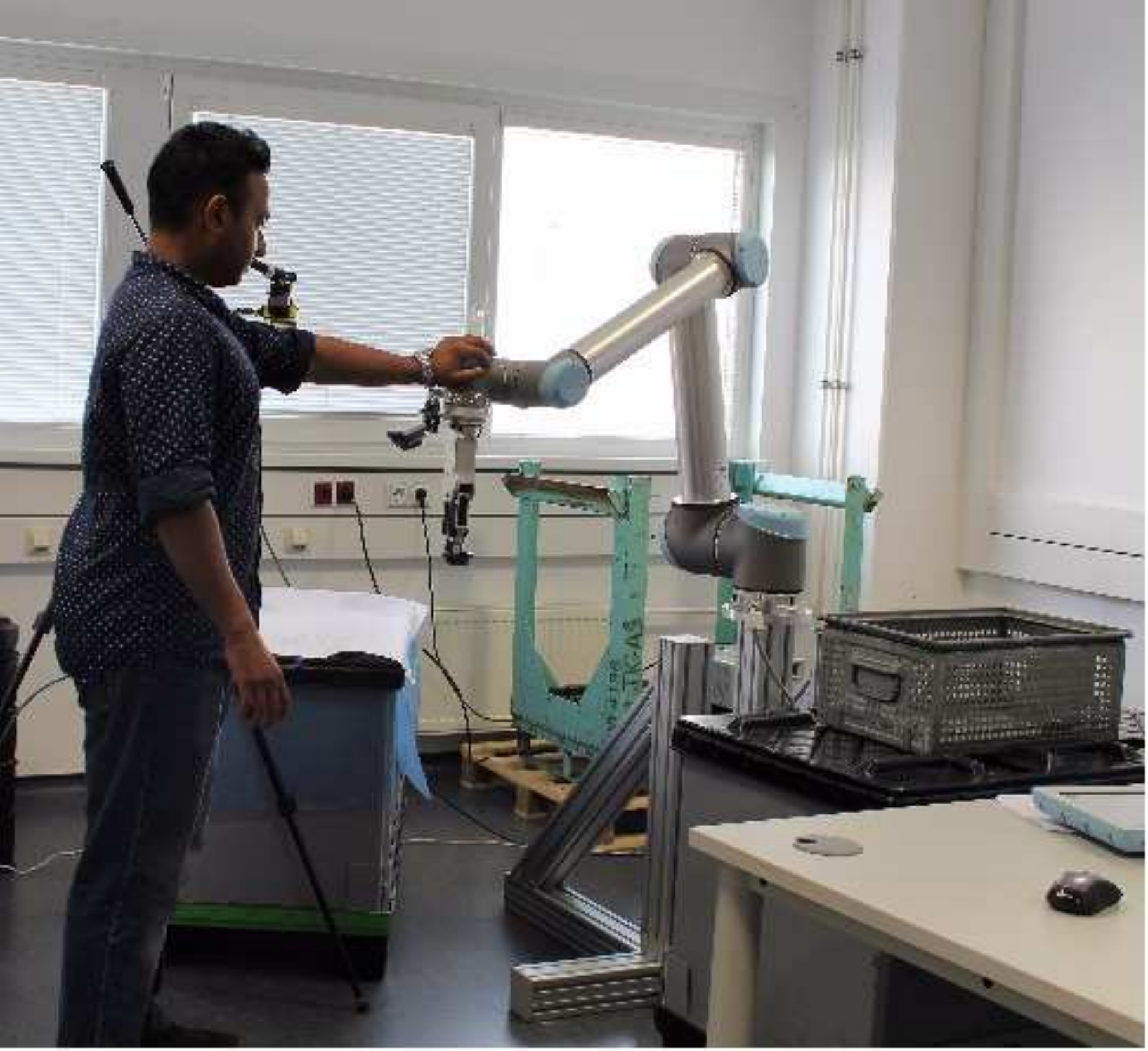}}}
\subfloat[]{
\resizebox*{4cm}{3cm}{\includegraphics{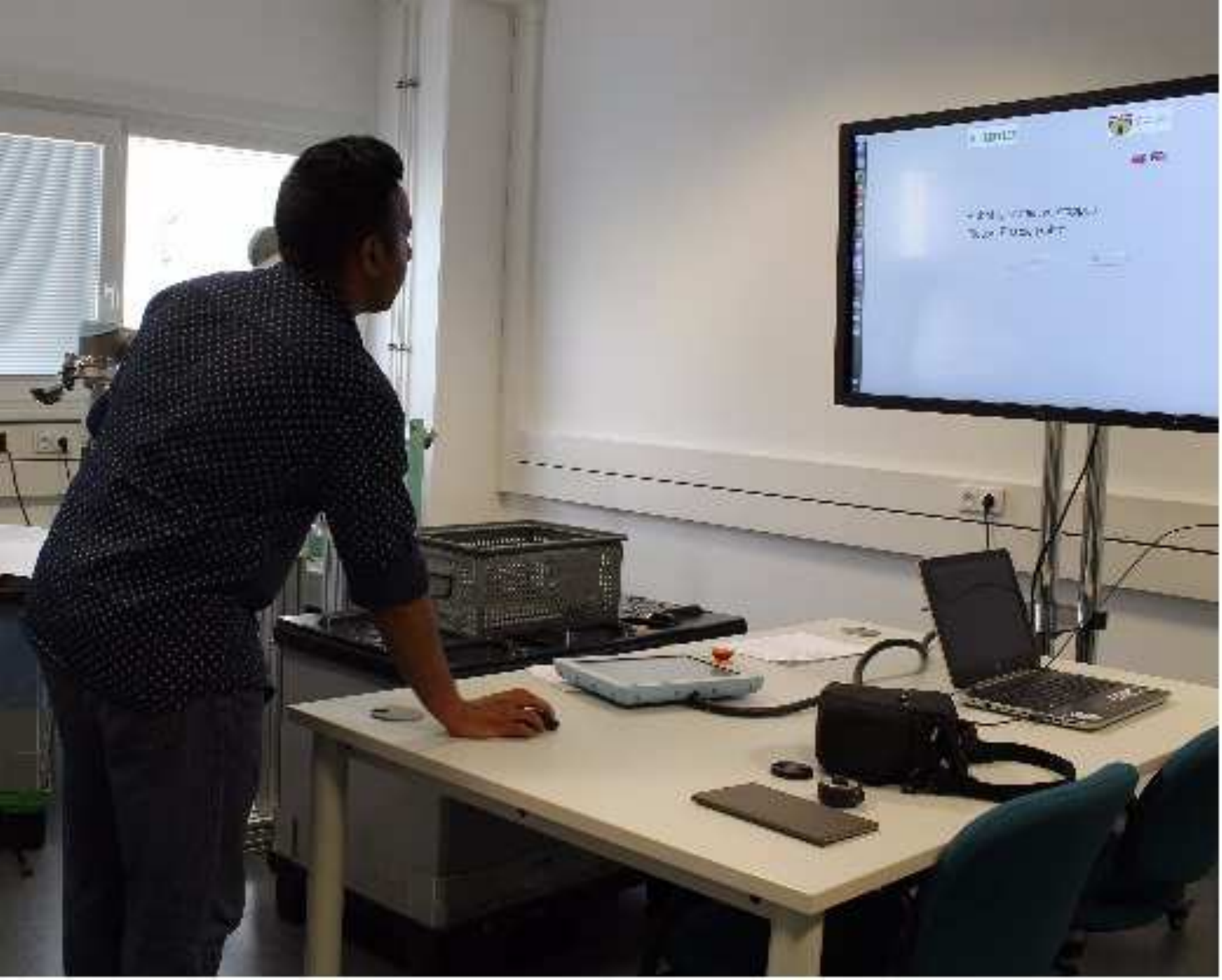}}}
\subfloat[]{
\resizebox*{4cm}{3cm}{\includegraphics{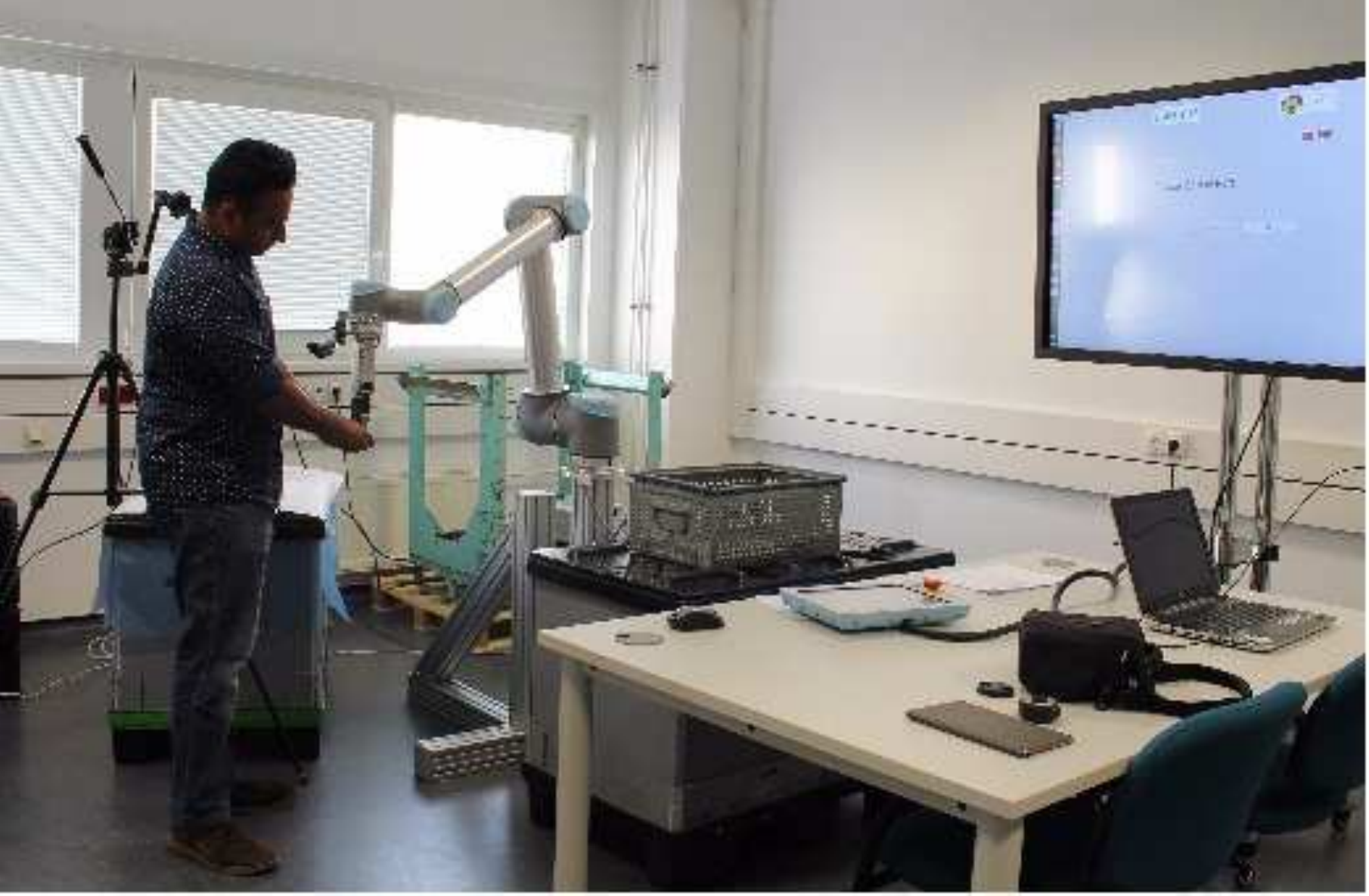}}}
\hfill
\subfloat[]{
\resizebox*{4cm}{3cm}{\includegraphics{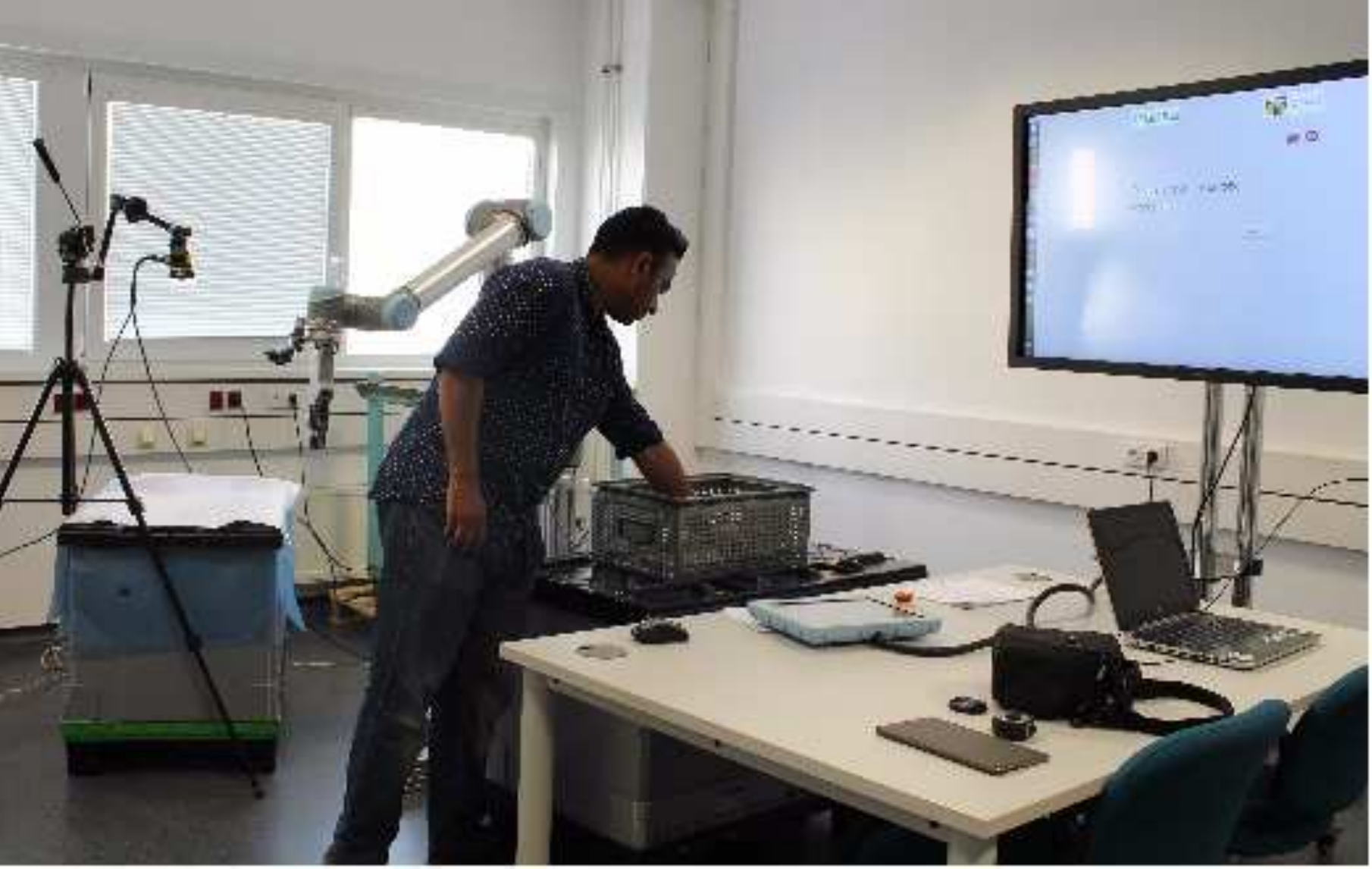}}}
\subfloat[]{
\resizebox*{4cm}{3cm}{\includegraphics{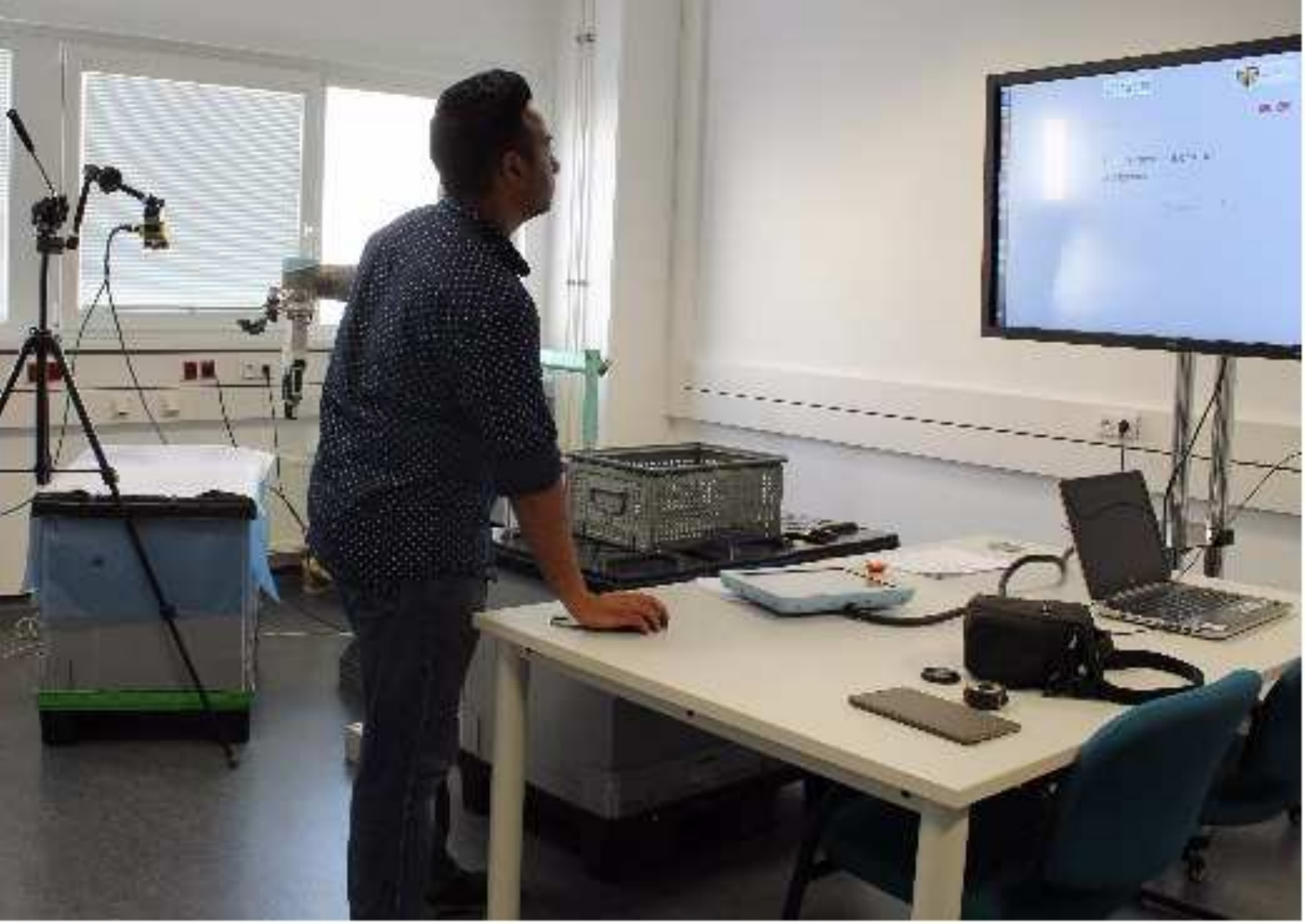}}}
\subfloat[]{
\resizebox*{4cm}{3cm}{\includegraphics{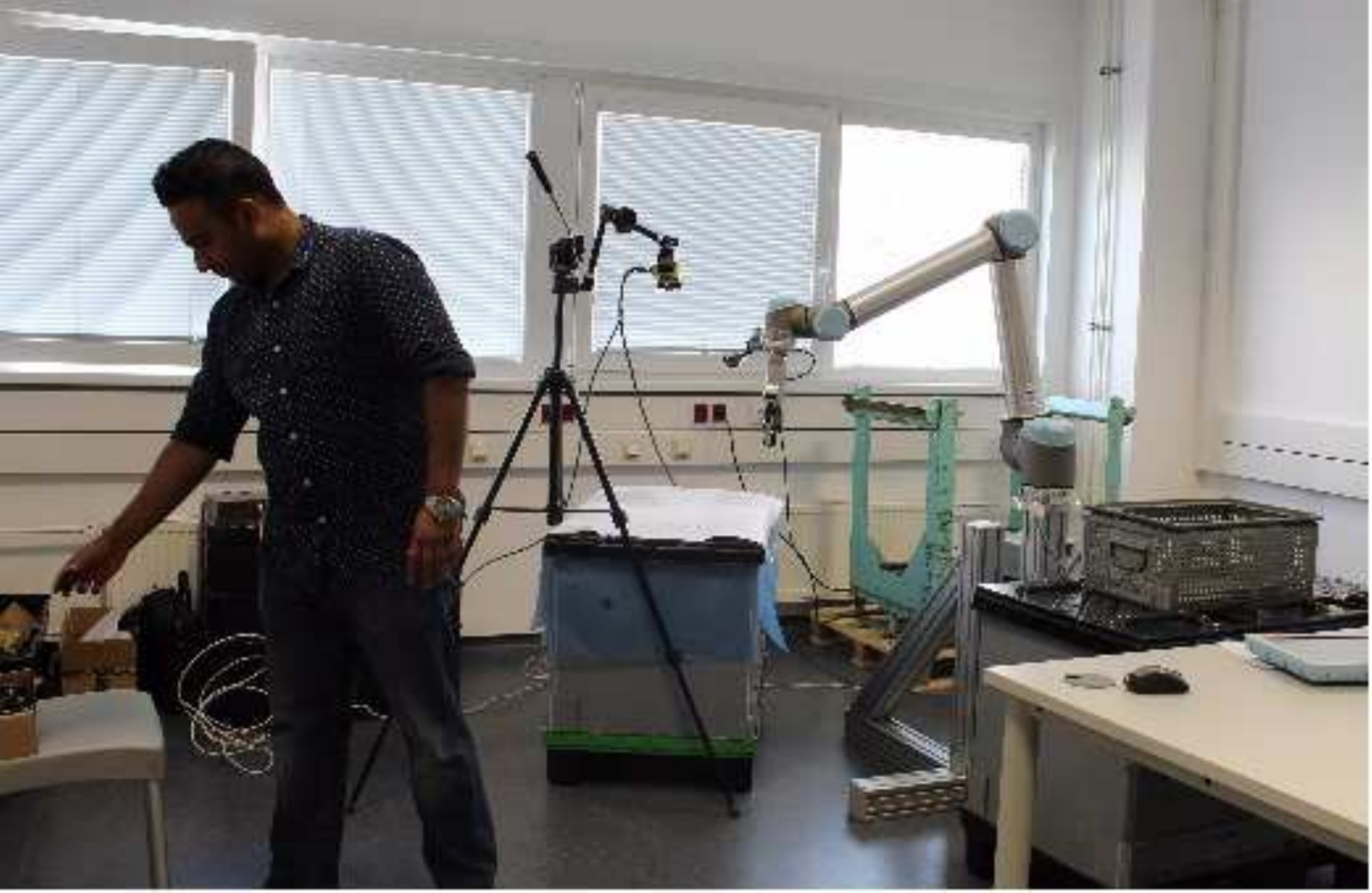}}}
\caption{An action sequence corresponding to the \textit{Collaboration} experiment: (a)-(c) initially proceeds along $h$ hyper-arc: the operator performs \texttt{inspect} and \texttt{deliver-part}, and the robot performs \texttt{approach-part} and \texttt{grasp} actions, (d) the operator stops the robot by exerting a force on its joints and the robot enters a protective stop, (e) the ongoing robot action is unsuccessful and \textit{Task Manager} switches to the next feasible cooperation path along the $hw$ hyper-arc. The feasible action is joint action \texttt{handover}. The operator interacts with \textit{User Interface} to enable the robot and commands the robot to release the part. (f) The robot opens the gripper and the operator retrieves the part (g) He places the part in pallet box (\texttt{palletize} action) and (h) he moves out of workspace and informs \textit{Task Manager} of action completion using UI. (i) Cooperation task continues with next action (\texttt{start-pose}).}
\label{fig:pickplace_new}
\end{center}
\end{figure*}

The three sub-experiments are detailed as follows:
\begin{enumerate}
\item
\label{experiment1}
\textit{Manual execution}.
Human operators take a part from the pickup box.
They perform visual and tactile inspection.
They place the part inside the pallet box in the correct sequence.
This is repeated until the pallet box is complete.
It can be observed that the robot is not used in this sub-experiment. 
The process is shown in Figure \ref{fig:manual_new}.
The weight of the part, the distance between the part and the pallet, and the pallet height causes physical fatigue and mental stress due to the monotonous nature of the activity.
\item
\label{experiment2}
\textit{Co-existence}.
Human operators pick up a part from the pickup box.
They perform visual and tactile inspection.
They place the part in the robot pickup area within the vision system's field of view.
Once the pose of the part is recognized, the robot grasps the part and places it inside the pallet.
The task is repeated for each part until the pallet is complete.
The process is detailed in Figure \ref{fig:handover_new}.
Additionally, we simulate a situation in which parts arrive at a faster rate, and as a consequence operators are forced to place multiple parts in the robot pickup area, simultaneously.
This shows the robustness of the architecture with the ability to handle moderate cluttered manipulation.
The process is shown in Figure \ref{fig:robust_new}.
In this case, there is task and workspace sharing between human operators and robots, however any contact is prevented during robot motion.
\item
\label{experiment3}
\textit{Collaboration}.
The robot executes the palletization task, and if the human operator enters the robot workspace (or the intervention area in Figure \ref{fig:tp-view}), and stops the robot by exerting a certain amount of force on the its body, then the robot immediately enters a protective stop when it detects a force of at least $100N$~\cite{URmanual}.
Human operators interact with the robot via the \textit{User Interface}, and the robot hands the part over to them.
The human operator places it in the correct position in the pallet box.
The robot continues with the next part.
The process is shown in Figure \ref{fig:pickplace_new}.
In addition to task and workspace sharing between the human operator and the robot, contact is allowed during robot motion.
This simulates the case when a random quality inspection needs to be performed, for instance by a manager.
It is necessary that the robot resumes smoothly from a protective stop without the need for the operator to access the robot controller. 
\end{enumerate}
Every trial was started with manual execution as the reference experiment.
This was followed by the co-existence experiment and the collaboration experiment.
A trial is considered to be successful when all three aforementioned sub-experiments are successful.
A trial is marked as unsuccessful if at least one sub-experiment fails.

Through the experiments we seek to evaluate the following hypotheses:
\begin{enumerate}
\item[$H_1$]
\label{hypo-stress}
\textit{Ergonomy}.
Human operators will feel less mentally stressed when using our architecture to collaborate with a robot than manually executing the task.
We will evaluate the mental stress through subjective measurements. 
\item[$H_2$]
\label{hypo-fluency}
\textit{Fluency}.
The architecture will lead to a natural and fluent interaction between the robot and a human operator.
For this hypothesis, we will use a combination of subjective and objective measures.
\item[$H_3$]
\label{hypo-flex}
\textit{Flexibility}.
The architecture will lead to an increased perception of flexibility in the HRC process for human operators.
For this hypothesis we will use subjective measurements.
\item[$H_4$]
\label{hypo-intimidation}
\textit{Intimidation}.
After the experiment, volunteers will feel less intimidated of working alongside a robot.
For this hypothesis, we will use subjective measurements through Likert-scale questionnaires before and after the experiment.
\end{enumerate}

\subsection{Objective Measures}
We compare the human operator action time, the robot action time and the time taken by the \textit{representation layer}, i.e., the \textit{Task Representation} and the \textit{Task Manager}, to complete the task.
In the following discussion, we assume a given sequence of $n$ actions ($a_1, a_2, \dots , a_n$).
We refer to $T_{m}$ as the time taken by the \textit{Task Manager} and the \textit{Task Representation} modules.
$T_{m}$ is defined as the sum of all such $n-1$ contributions (the first being set by default on the optimal path), where each contribution is given by the difference between the time $T_{next}$ when the next action suggestion for $a_i$, with $i \geq 2$ and $i \leq n$, is ready and the time $T_{ack}$ when an acknowledge for a previous action $a_{i-1}$ is received by the \textit{Task Manager}, such that:
\begin{equation}
T_{m} = \sum_{i=2}^{n}T_{next}(a_i) - T_{ack}(a_{i-1}).
\end{equation}
We refer to $T_h$ as the amount of time a human operator takes to perform all actions $a_{i}$ in the sequence. 
In our experiments, $T_h$ is computed as the time taken by the perception layer to recognize the object pose.
This is due to the fact that we do not actively measure or track actions carried out by human operators.
This quantity is computed in the \textit{Human Interface} module as:
\begin{equation}
    T_{h} = \sum_{i=1}^{n}T_{resp}(a_i) - T_{recv}(a_i),
\end{equation}
where $T_{recv}$ is the time instant when \textit{Human Interface} receives a command, and $T_{resp}$ is the time instant when the \textit{Human Interface} sends back the response.
It is noteworthy that this measures overestimate the actual contribution by a human operator.
Similarly, $T_r$ refers to the amount of time the cobot is actually active, i.e., when the robot is performing either a motion or a manipulation task.
$T_r$ is computed by the \textit{Robot Interface} module as:
\begin{equation}
    T_{r} = \sum_{i=1}^{n}T_{resp}(a_i) - T_{recv}(a_i),
\end{equation}
where in this case $T_{recv}$ is the time instant when the \textit{Robot Interface} receives a command, and $T_{resp}$ is the time instant when the \textit{Robot Interface} sends back the response.
Average time and standard deviation values for $T_m$, $T_h$ and $T_r$ are reported in Table \ref{table:coop-time}.
We also refer to $T_c$ as the total cooperation time, which is related to the successful execution of the whole experiment.
Furthermore, Table \ref{table:hrcComparison} compares the \textit{Manual execution} experiment with the experiments when the robot is used, on metrics such as average completion time, number of tasks assigned to human operators or robots, and the weight handled.
Results are analysed in Section \ref{Experiment:discussion}.

\begin{table*}
\centering
\caption{Total cooperation time, and average time required by human operators, the robot, and their standard deviations for successful experiments.}
\label{table:coop-time}
\begin{tabular}{|c|c|c|c|c|}
\hline
\textit{Measure}			& \textit{Average time (s)} 		& \textit{Percentage of total time (\%)} 	& \textit{Standard deviation (s)} 	& \textit{Standard deviation (\%)} 	\\ \hline \hline
$T_{m}$				& 2.49 					& 0.51 							& 0.16 						& 0.04 						\\ 
$T_h$				& 77.75 					& 17.97 							& 19.20 						& 3.72 						\\ 
$T_r$    				& 401.57 					& 81.52 							& 10.19 						& 3.51 						\\ 
$T_c$ 				& 481.82 					& 100 							& 24.34 						&  							\\ \hline
\end{tabular}
\end{table*}

\subsection{Subjective Measures}
\begin{figure*}[t!]
\begin{center}
\subfloat[]{
\resizebox*{0.5\textwidth}{!}{\includegraphics{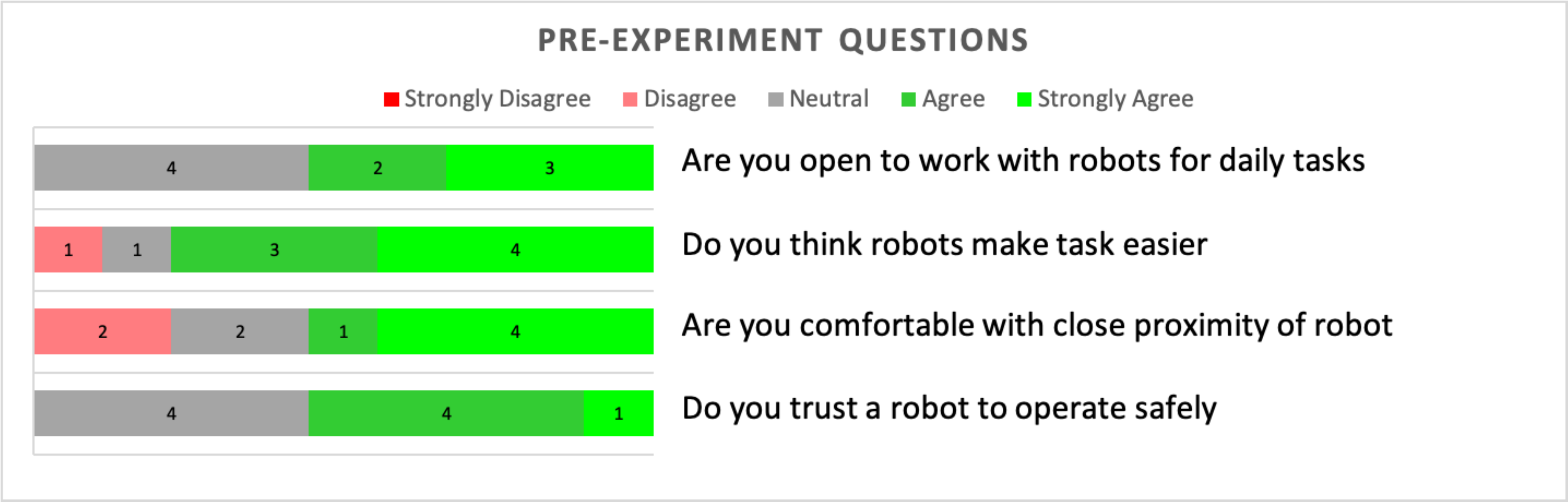}}}
\subfloat[]{
\resizebox*{0.5\textwidth}{!}{\includegraphics{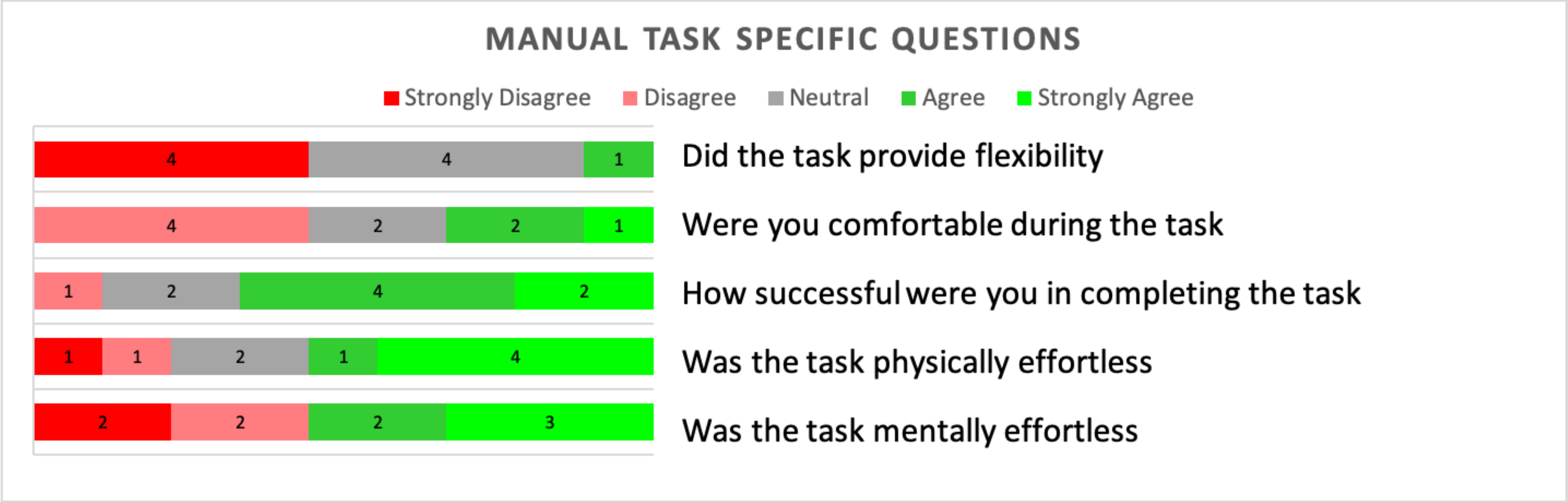}}}\hfill
\subfloat[]{
\resizebox*{0.5\textwidth}{!}{\includegraphics{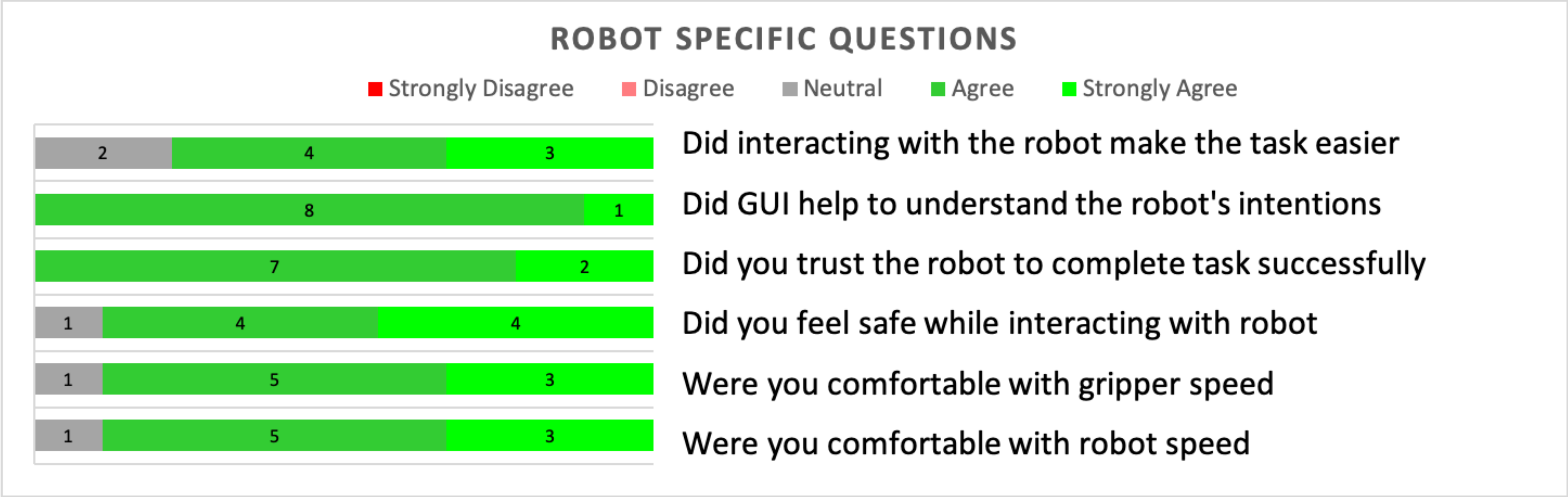}}}
\subfloat[]{
\resizebox*{0.5\textwidth}{!}{\includegraphics{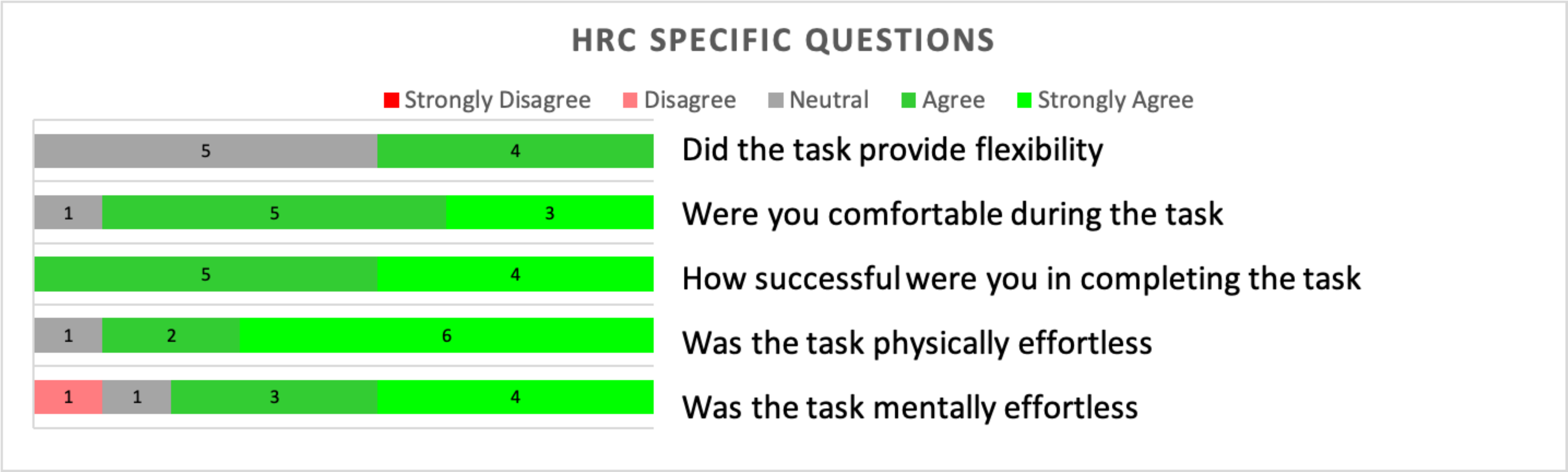}}}
\caption{Results of the Likert-scale questionnaire survey conducted at Schaeffler with 9 volunteers: (a) pre-experiment questions, (b) manual task specific questions (c) robot-specific questions (d) HRC task-specific questions.}
\label{fig:survey}
\end{center}
\end{figure*}

Volunteers were asked to fill a set of Likert-scale questions before and after each experiment to understand their outlook towards robots as co-workers. 
Questions are shown in Figure \ref{fig:survey}.
Replies assume a numerical value of $1$ for \textit{strongly disagree}, and $5$ for \textit{strongly agree}.
Questions labelled as \textit{pre-experiment questions} were asked during the familiarization phase, prior to any experiment, whereas the questions termed \textit{manual task-specific questions} were asked after they finished performing the manual task experiment during the test phase.
Similarly, after the \textit{Co-existence} and \textit{Collaboration} experiments, volunteers were asked a set of \textit{robot-specific questions} and \textit{HRC-specific questions}.
The latter questionnaire provides a subjective evaluation of the quality of human-robot collaboration similar to that used in~\cite{shah2011fluid}: 
volunteers rate their agreement to statements related to their trust in the robot, robot's performance, team fluency, mental effort during collaboration, and comfort level when in close proximity with the robot.
Results are analysed in Section \ref{Experiment:discussion}.

\subsection{Results and Discussion}
\label{Experiment:discussion}
This Section discusses the outcomes of the experiments from both a qualitative and quantitative perspective. 
During the experimental trials, there were $6$ failed experiments out of a total number of $30$ (80\% success rate). 
The reasons for the failures are: 
(i) failure in properly picking-up the part due to incorrect grasping pose recognition by the \textit{Object Pose Recognition} module (in four cases),
(ii) the robot dropping the part during transportation due to improper grasping (in one case), and 
(iii) human error during the \texttt{deliver-part} action, i.e., the part was accidentally moved after the vision system determined the pose parameters for the robot (in one case).
Hence, failures cannot be attributed to the \textit{Task Representation} or to the \textit{Task Manager} modules, but rather due to a lack of robustness in the employed vision and manipulation algorithms, which are not the main focus of this paper. 
Considering the objective measures from Table \ref{table:coop-time}, we first observe that the standard deviation associated with $T_m$ is low. 
This means that the related time interval takes almost identical time duration for each trial, and demonstrates the temporal deterministic nature of \textit{Task Manager} and \textit{Task Representation}.
Such a temporal determinism associated with the \textit{representation layer} is an advantageous feature of the proposed framework.
In contrast, the standard deviation is higher when considering the time required by the human operator or the robot activities.
This is understandable, as human operators take different amounts of time to perform each action, and the robot action time also varies whenever there is human intervention.
Furthermore, the only time an agent (either human operator or robot) is not performing an action is when the \textit{Task Manager} is active, which takes only about 0.54\% of the entire cooperation time.
As a consequence, we can argue that the overall idle time is minimised, which is another desirable feature of the framework.
From these measurements, we can conclude that the system ensures fluency in human-robot collaboration, thereby corroborating hypothesis $H_2$.

\begin{table*}
\centering
\caption{Comparison between \textit{Manual execution} versus \textit{Co-existence} and \textit{Collaboration} (collapsed).}
\label{table:hrcComparison}
\begin{tabular}{|c|c|c|}
\hline
\textit{Property}					& \textit{Manual execution} 	& \textit{Co-existence} and \textit{Collaboration}		\\ \hline \hline
$T_c$ (min)					& 3.15					& 8.03                               						\\ 
$|a|_h$						& 7						& 5										\\ 
$|a|_r$						& 0						& 2										\\ 
$L_h$ (kg)					& 0.1-5.5              			& Negligible 								\\ 
$L_r$ (kg)						& 0						& 0.1-5.5									\\ \hline
\end{tabular}
\end{table*}

\begin{table*}
\centering
\caption{ Result of the inferential statistics analysis highlighting the difference between the manual and HRC execution of the task. }
\label{table:statistics}
\begin{tabular}{|c|c|c|c|c|c|}
\hline
\textit{Studied feature} & \#1 \textit{Flexibility} & \#2 \textit{Comfort} & \#3 \textit{Success} & \#4 \textit{Physically effortless}	& \#5 \textit{Mentally effortless}	\\ \hline \hline 
\textit{p-value} & 0.010 & 0.016 & .050 & 0.069	& 0.169
\\ \hline
\end{tabular}
\end{table*}

\begin{table*}
\centering
\caption{Comparison between our proposal and a few other state-of-art approaches to solve assembly tasks using human-robot collaboration.}
\label{table:tabcomparison}
\resizebox{\textwidth}{!}{%
\begin{tabular}{|l|l|l|l|l|l|l|l|}
\hline
                             & \multicolumn{1}{c|}{\textbf{Task representation}}                    & \multicolumn{1}{c|}{\textbf{\begin{tabular}[c]{@{}c@{}}Architecture \\ type\end{tabular}}} & \multicolumn{1}{c|}{\textbf{Task allocation}} & \multicolumn{1}{c|}{\textbf{Sensors}} & \multicolumn{1}{c|}{\textbf{\begin{tabular}[c]{@{}c@{}}Sound and \\ Complete\end{tabular}}} & \multicolumn{1}{c|}{\textbf{Flexibility}} & \multicolumn{1}{c|}{\textbf{Adaptability}} \\ \hline \hline 
                             
\textbf{Our Proposal}        & AND/OR graph                                                         & Online                                                                                     & Offline                                       & RGB camera                            & Yes                                                                                        & High                                      & Yes                                        \\ \hline
\textbf{Johannsmeier et al. \cite{johannsmeier2017hierarchical}} & AND/OR graph                                                         & Offline                                                                                    & Offline                                       & N.A.                                  & Yes                                                                                         & Low                                       & No                                         \\ \hline
\textbf{Hawkins et al. \cite{hawkins2014anticipating}}      & AND/OR graph                                                         & Online                                                                                     & Offline                                       & RGB-D                                 & N.A.                                                                                        & High                                    & Yes                                        \\ \hline
\textbf{Lamon et al. \cite{lamon2019capability}}        & AND/OR graph                                                         & Offline                                                                                    & Offline                                       & Microsoft HoloLens                    & Yes                                                                                         & None                                      & No                                         \\ \hline
\textbf{Darvish et al. \cite{darvish2018interleaved}}      & AND/OR graph                                                         & Online                                                                                     & Online                                        & RGB-D+wearable sensor                                 & Yes                                                                                        & High                                      & Yes                                        \\ \hline
\textbf{Tsarouchi et al. \cite{tsarouchi2017human}}    & Scheduling problem                                                   & Offline                                                                                    & Online                                        & Depth sensor                          & Yes                                                                                        & None                                      & No                                         \\ \hline
\textbf{Wilcox et al. \cite{wilcox2013optimization}}       & Scheduling problem                                                   & Online                                                                                     & Offline                                       & None                                  & Yes                                                                                         & High                                      & Yes                                        \\ \hline
\textbf{Paxton et al. \cite{paxton2017costar}}       & Behavior Trees                                                       & Offline                                                                                    & Offline                                       & RGB-D                                 & \begin{tabular}[c]{@{}l@{}}Sound but not\\  complete\end{tabular}                           & None                                      & No                                         \\ \hline
\textbf{Toussaint et al. \cite{toussaint2016relational}}    & \begin{tabular}[c]{@{}l@{}}Markov Decision \\ Processes\end{tabular} & Online                                                                                     & Offline                                       & Motion capture                        & Yes                                                                                         & N.A.                                    & Yes                                        \\ \hline
\end{tabular}%
}
\end{table*}

Table \ref{table:hrcComparison} shows the comparison between the \textit{Manual execution} case and the collapsed \textit{Co-existence} and \textit{Collaboration} cases. 
The total amount of weight $L_h$ (kg) carried by human operators is negligible in the case of HRC-related cases, in comparison to the \textit{Manual execution} case in which operators handle a maximum weight of around $5.5$ $kg$.
This is in line with requirement \ref{obj1} dealing with ergonomy.
Considering the requirement \ref{obj3}, which is related to computational performance, it can be seen from Table \ref{table:hrcComparison} that the amount of time $T_c$ taken to complete the task while collaborating with a robot almost doubles the time taken for a human operator to complete the process alone.
This can originate from different causes.
Firstly, the maximum allowed velocity of the robot's end-effector is set at $250$ $mm/s$ according to ISO 10218 standards, hence robot motions consume around 80\% of the total time. 
For \textit{Co-existence} tasks, with the availability of sensors allowing for an intelligent workspace sharing~\cite{robla2017working}, the robot velocity can be increased thus reducing the overall cooperation time.
Secondly, a normal $8$-hour work shift induces fatigue and possible lapse in concentration in human operators, due to the monotonous nature of the task. 
However, it is not possible to simulate the exact psychological conditions during the manual task execution experiment, which lasts for a short period of time.
Additionally, since the experiment was video-taped and monitored by an audience, volunteers were motivated to complete the task as effectively as possible, which is also testified by comparing the number of actions $|a|_h$ and $|a|_r$ carried out by the human operators or the robot, respectively, in the two cases.

Regarding the subjective measure from Figure \ref{fig:survey}, we present the results of Likert-scale questionnaires presented to 9 volunteers since the results from one questionnaire are unavailable.
Prior to the experiments we found that volunteers who had no prior experience with robots were more likely to be intimidated by the robot working in close proximity. 
Subsequently, eight out of nine volunteers felt safe while interacting with the robot after the experiment, including those who initially felt uncomfortable with working next to the robot.
This result supports Hypothesis $H_4$ regarding intimidation while working close to a robot without any safety barriers.
Furthermore, there is a correlation with the objective measures from Table \ref{table:hrcComparison} that negligible weight handled by human operators leads to a reduced physical stress. 
Additionally, there was consensus that interacting with the \textit{User Interface} module helped volunteers understand robot intentions.
Most human operators felt confident with the physical contact with a moving robot.
This is noteworthy, since during an intervention process, a human operator stops the moving robot and waits for the gripper to open.
Using collaborative grippers with soft fingertips was justified, which may not have been the case if we had used traditional high-powered pneumatic grippers with metal fingers.
Given the low number of subjects in this study, we conducted a T-test as a statistical analysis of our hypothesis. 
The alternative hypotheses are followed by the questionnaires (b) and (d) shown in Figure \ref{fig:survey}, whereby human-robot collaboration compared to the manual execution of the task enhances the flexibility, comfort, and success as well as a reduction in the physical and mental effort of workers. 
The T-scores are identified on the basis of a two-tailed paired (correlated) T-test, considering that the individual subjects are examined for different test scenarios. 
Accordingly, the rounded p-values are reported in Table \ref{table:statistics}. Prior to the experiments, the significance level is selected as $\alpha = 0.05$. These evidence strongly rejects the null hypothesis concerning the HRC flexibility and comfort of the user, therefore highly endorsing our claim regarding the flexibility ($H_3$) and comfort (partially $H_1$) of the HRC system. 
However, evidence does not support the enhancement of the user workload, i.e., the mental and physical efforts, and the perceived success of the approach is not of statistical significance as well.

We compare our proposal with other state-of-art approaches for assembly tasks involving human-robot collaboration in Table \ref{table:tabcomparison}. 
The metrics used for comparison are the method for task representation, architecture type (online or offline), task allocation (online or offline), the adopted sensors, soundness and completeness, flexibility, and adaptability. 
For the sake of clarity, we describe the metrics used for this comparison. 
The architecture is classified as offline if the cooperation model or path is pre-determined before the cooperation process, whereas, if it is computed during the cooperation process, then the architecture type is tagged online. 
Similarly, task allocation can be classified as offline or online whether the allocation process is predetermined or not, respectively. 
The architecture is said to be \textit{sound} if a provided solution is guaranteed to be true. 
Conversely, the architecture is said to be \textit{complete} if it is guaranteed to return a solution if it exists. 
The architecture is flexible if the involved agents have the freedom to choose which action to perform during the interaction. 
This further entails the ability of the robot to react to varying human actions online. 
We classify the various approaches based on the flexibility dimension with the levels none, low, and high. 
This is due to the fact that various approaches define flexibility differently and we assign the corresponding value based on the evidence provided in the experiments. 
Similarly, adaptability is the ability of the architecture to showcase online re-planning in case of varying or unforeseen human action. 
Most of the approaches in the Table \ref{table:tabcomparison} use AND/OR graphs such as our proposal, i.e., \cite{johannsmeier2017hierarchical}, \cite{hawkins2014anticipating}, \cite{lamon2019capability}, and \cite{darvish2018interleaved}, while other popular techniques include scheduling problems (\cite{tsarouchi2017human}, \cite{wilcox2013optimization}), BTs \cite{paxton2017costar}, and Markov Decision Processes (MDPs) \cite{toussaint2016relational}. 
The set of online architectures include our proposal, \cite{hawkins2014anticipating}, \cite{darvish2018interleaved}, \cite{wilcox2013optimization}, and \cite{toussaint2016relational}, whereas the others work offline. 
With the exception of \cite{darvish2018interleaved} and \cite{tsarouchi2017human}, which offer online task allocation, the task allocation process in all other approaches is done offline. 
Sensors commonly used are RGB or RGB-D cameras, while few techniques propose the use of wearable devices \cite{darvish2018interleaved} and mixed reality technology \cite{lamon2019capability} for solving the perception and human action recognition problem. 
All the approaches are sound and complete with the exception of \cite{paxton2017costar}, where the use of BTs entail the architecture to be sound but not complete. Furthermore, in terms of flexibility and adaptability, approaches based on our proposal, \cite{hawkins2014anticipating}, \cite{darvish2018interleaved}, \cite{wilcox2013optimization} provide high flexibility and adaptability. 
In contrast, for what concerns the method based on MDPs to solve a reinforcement learning problem, described in \cite{toussaint2016relational}, while ensuring the presence of an adaptable architecture, it is unclear on the level of flexibility provided. 
For what regards the work in \cite{johannsmeier2017hierarchical}, although the authors claim their proposal to be flexible, to the best of our understanding no conclusive evidence for high flexibility or adaptability is provided in the paper. 
Similarly, techniques such those described in \cite{lamon2019capability}, \cite{tsarouchi2017human} and \cite{paxton2017costar}, do not emphasise the availability of a flexible and adaptable architecture according to the definitions stated above.    
From Table \ref{table:tabcomparison}, we can reinforce our claim that our proposal provides high flexibility for the humans during the cooperation process while also being comparable to other state-of-art approaches for the aforementioned parameters. 
In addition, we also employ an informative user interface, which allows for a sub-linguistic communication level between the agents and helps the workers understand robot's intentions.
Furthermore, its noteworthy that our approach has been validated in a real industrial setting with factory line workers.

The proposed architecture is also characterised by a few limitations, which are to be considered for future research activities.
\begin{enumerate}
\item 
Cooperation models take the form of turn-taking between agents (either robots or human operators), with a few implicit, non-modelled, turns where physical collaboration is necessary.
However, physical human-robot collaboration can increase the scope of the architecture to a wide variety of more realistic shop-floor activities.
\item
Task allocation in our framework is performed \textit{a priori}, either to the robot or to human operators, depending on their capabilities. 
Obviously enough, on-the-fly allocations would increase the flexibility of the whole cooperation process, also in view of robot or human capabilities \cite{Mastrogiovannietal2013}
\item 
Robot actions are deterministic, i.e., we assume that no errors can be managed. 
This has important consequences on the cooperation tasks. 
As a matter of fact, increasing the robustness of robot manipulation through, e.g., \textit{visual servoing} \cite{espiau1992new}, or incorporating additional sensors to actively monitor the shared workspace can greatly improve the effectiveness of the cooperation task~\cite{michalos2015design}. 
\end{enumerate}

\section{Conclusions}
\label{sec:conclusion}
This paper describes a hybrid, reactive-deliberative architecture for human-robot collaboration in industrial scenarios. 
The architecture is characterised by two interesting advantages with respect to existing state-of-the-art approaches.
First, and foremost, the architecture allows a collaborative robot to plan for an optimal sequence of actions, to be carried out either by the robot or a human operator, and to reactively adapt to unforeseen human actions on line, thereby allowing for a natural and intuitive interaction.
Second, the framework employs a sub-linguistic, semantic-oriented, communication level, whose related information is grounded on the underlying task representation mechanism.
Finally, the results have been compared with other works previously proposed in the literature, and a statistical analysis according to the feedback of production line workers has been carried out.

\section*{Acknowledgements}
The authors would like to thank the Schaeffler Group, Slovakia, for the opportunity to perform research in their manufacturing plant, and solve real-world problems in human-robot collaboration.
The work has been supported by the European Master on Advanced Robotics Plus (EMARO+) programme.

\bibliographystyle{Classes/tADR}
\bibliography{Classes/Bibliography}
\end{document}